\documentclass{article}

 \usepackage[main, final]{neurips_2026}

\usepackage[utf8]{inputenc} 
\usepackage[T1]{fontenc}    
\usepackage{hyperref}       
\usepackage{url}            
\usepackage{booktabs}       
\usepackage{amsfonts}       
\usepackage{nicefrac}       
\usepackage{microtype}      
\usepackage{xcolor}         
\usepackage{amsmath}
\usepackage{amssymb}    
\usepackage{cleveref}   
\usepackage{xspace}     
\usepackage{algorithm}
\usepackage{algorithmic}

\usepackage[most]{tcolorbox}
\usepackage{colortbl}
\usepackage{caption}
\usepackage{multirow}
\usepackage{graphicx} 

\usepackage{subcaption}
\usepackage{wrapfig}
\usepackage{enumitem}

\newcommand{\sys}{\textsc{ECHO-2}\xspace}
\newcommand{\sysname}{\textsc{ECHO-2}\xspace}
\newcommand{\verl}{\textsc{verl}\xspace}

\definecolor{citepink}{HTML}{8470FF}

\hypersetup{
    colorlinks=true,
    citecolor=citepink,  
    linkcolor=red,     
    urlcolor=black,      
    filecolor=black
}

\tcbuselibrary{listings, skins} 

\newtcblisting{codebox}[1][]{
    enhanced, 
    colback=white,       
    colframe=black,      
    boxrule=0.4pt,       
    arc=0pt,             
    outer arc=0pt,
    shadow={1.5mm}{-1.5mm}{0pt}{blue!10!white}, 
    listing only,
    listing options={
        language=Python,             
        basicstyle=\ttfamily\small,
        keywordstyle=\color{magenta}, 
        numbers=left,              
        numberstyle=\tiny\color{gray}, 
        numbersep=8pt,             
        breaklines=true,           
        escapechar=|,              
    },
    left=6mm,
    top=0mm,
    bottom=0mm,
    #1 
}

\newtcolorbox{promptbox}[1][]{
  enhanced,
  breakable,
  colback=promptboxlightgray,
  colframe=promptboxblue!30,
  arc=8pt,
  boxrule=0.5pt,
  left=12pt,
  right=12pt,
  top=8pt,
  bottom=8pt,
  fonttitle=\bfseries,
  fontupper=\linespread{1.2}\selectfont,
  title=#1
}

\title{\sys: A Large-Scale Distributed Rollout Framework for Cost-Efficient Reinforcement Learning}

\author{%
Jingwei Song$^{1,3*}$,
Meng Chen$^{2,*}$,
Jie Xiao$^{3,*}$,
Qingnan Ren$^{3,*}$,
Jiaqi Huang$^{3}$, \\
\textbf{Yangshen Deng}$^{4}$,
\textbf{Chris Tong}$^{3}$,
\textbf{Wanyi Chen}$^{5}$,
\textbf{Suli Wang}$^{6}$,
\textbf{Zhisheng Chen}$^7$, \\
\textbf{Ziqian Bi}$^{3}$, 
\textbf{Shuo Lu}$^{3}$,
\textbf{Yiqun Duan}$^{3}$,
\textbf{Xu Wang}$^{3}$,
\textbf{Rymon Yu}$^{3}$, \\
\textbf{Ween Yang}$^{3}$, 
\textbf{Lynn Ai}$^{3}$,
\textbf{Eric Yang}$^{3}$,
\textbf{Tianyu Shi}$^{8,\dagger}$ \\
$^{1}$The University of Hong Kong,
$^{2}$Fudan University,
$^{3}$Gradient,
$^{4}$University of Edinburgh, \\
$^{5}$Soochow University,
$^{6}$Technical University of Darmstadt, \\
$^{7}$University of the Chinese Academy of Sciences,
$^{8}$McGill University\\
\href{mailto:u3638265@connect.hku.hk}{\texttt{songjingwei@connect.hku.hk}},   
\href{mailto:tianyu@gradient.network}{\texttt{tianyu.shi3@mcgill.ca}} \\
{\small $^*$Equal contribution, $^\dagger$Corresponding author}
}

\begin{document}
\maketitle

\begin{abstract}
Reinforcement learning (RL) is a critical stage in post-training large language models (LLMs), involving repeated interaction between rollout generation, reward evaluation, and centralized learning. Distributing rollout execution offers opportunities to leverage more cost-efficient inference resources, but introduces challenges in wide-area coordination and policy dissemination. We present ECHO-2, a distributed RL framework for post-training with remote inference workers and non-negligible dissemination latency. ECHO-2 combines centralized learning with distributed rollouts and treats bounded policy staleness as a user-controlled parameter, enabling rollout generation, dissemination, and training to overlap. We introduce an overlap-based capacity model that relates training time, dissemination latency, and rollout throughput, yielding a practical provisioning rule for sustaining learner utilization. To mitigate dissemination bottlenecks and lower cost, ECHO-2 employs peer-assisted pipelined broadcast and cost-aware activation of heterogeneous workers. Experiments on GRPO post-training of LLMs ranging from 4B to 32B parameters under real wide-area bandwidth regimes show that ECHO-2 significantly improves cost efficiency while preserving RL reward comparable to strong baselines.
\end{abstract}

\section{Introduction}

Reinforcement learning (RL) has become a central component of post-training for large language models (LLMs), driving improvements in reasoning, tool use, safety alignment, and preference optimization at scale~\citep{team2025kimi,guo2025deepseek,shao2024deepseekmath}. While modern algorithms such as PPO~\citep{schulman2017proximalpolicyoptimizationalgorithms} and GRPO~\citep{shao2024deepseekmath} have advanced stability and sample efficiency, the system design of LLM RL remains largely conventional: most pipelines co-locate learners and rollout workers inside a data center and run tightly coupled iteration cycles~\citep{sheng2025hybridflow}.

This design conflicts with the cost structure of contemporary RL workloads. Rollout generation dominates wall-clock time and often leaves the learner intermittently idle~\citep{sheng2025hybridflow,fu2025areal,he2025history}, yet rollouts---which consist primarily of forward passes and reward evaluation---are executed on the same expensive GPU clusters used for learning. Recent asynchronous systems~\citep{noukhovitch2024asynchronous,fu2025areal,he2025history,zhong2025streamrl,xiao2025echo} relax strict synchronization to improve utilization, but still assume rollout workers and learners share an administrative domain with provisioned interconnects~\citep{zhong2025streamrl}; they do not change the underlying cost structure. Distributed inference resources, in contrast, are abundant and cheap---geographically distributed cloud instances or opportunistic compute~\citep{borzunov2023petals,ryabinin2023swarm}---and RL training is agnostic to where trajectories originate. However, naively extending centralized designs to such settings yields severe inefficiencies (training bubbles, overprovisioning) due to heterogeneous throughput, wide-area latency, and dynamic availability.

This motivates our research question: \emph{how can we reduce the cost of RL post-training by offloading rollout generation to distributed inference resources, while keeping a centralized learner continuously utilized?}

We present \sysname{}, a distributed RL framework built on a simple architectural principle: \emph{centralized learning with distributed rollouts.} Policy optimization runs on a small, stable set of data-center GPUs, while rollout generation is offloaded to a heterogeneous pool of Parallax~\citep{tong2025parallax} inference workers connected over wide-area networks. \sysname{} achieves cost-efficient operation through two complementary mechanisms. First, it adopts a \textbf{bounded-staleness execution model}~\citep{fu2025areal,he2025history,zhou2025rlax}: the learner consumes rollouts whose policy lags by at most $S$ training steps, where $S$ is a user-specified staleness budget. This temporal slack absorbs wide-area latency and lets rollout generation, dissemination, and training overlap without stalling the learner. Second, \sysname{} treats \textbf{policy dissemination} as an engineered primitive: workers organized into peer-forwarding chains immediately relay newly received snapshots and begin generating rollouts as soon as installation completes, leveraging aggregate fleet bandwidth to reduce tail broadcast latency. Together, these mechanisms turn $S$ from an artifact to be minimized into a \emph{system-level control knob} that trades rollout cost against training stability.

This perspective reduces a hard synchronization constraint to a provisioning problem: given $S$, how much aggregate rollout capacity is required to saturate the learner? \sysname{} answers this with a closed-form rule relating measurable training time, dissemination latency, and per-worker throughput to the required capacity. Evaluated on GRPO post-training of 4B and 8B models across distributed rollout pools and wide-area bandwidth regimes, \sysname{} substantially reduces end-to-end training cost while matching the RL quality of strong centralized baselines.

\noindent\textbf{In summary, our contributions are:}
\begin{itemize}[
    leftmargin=1.2em,
    labelsep=0.45em,
    itemsep=2pt,
    topsep=3pt,
    parsep=0pt,
    partopsep=0pt
]
  \item \textbf{A distributed inference RL architecture for cost-efficient post-training.}
  We propose a system architecture that separates centralized learning from
  distributed rollout inference, enabling RL post-training to reduce cost by
  offloading rollout generation from data-center GPU clusters to distributed resources.

  \item \textbf{Overlap-aware execution and peer-assisted broadcast.}
  We design system mechanisms that enable overlapping rollout inference, policy dissemination, and training across distributed rollout workers and a centralized trainer via a simple provisioning rule. \sysname{} bounds policy staleness by a user-specified budget $S$ and employs peer-assisted broadcast to reduce dissemination tail latency.

  \item \textbf{Three-plane disaggregation of rollout, learning, and data.}
  \sysname{} cleanly decouples rollout inference, policy optimization, and
  data handling into independent execution planes, enabling flexible integration of new RL tasks.

  \item \textbf{End-to-end evaluation on LLM RL workloads.}
  Through extensive end-to-end experiments, we show that \sysname{} substantially reduces the cost of RL post-training while maintaining learning quality, making large-scale RL more accessible under realistic resource constraints.
\end{itemize}

\begin{figure}[t]
    \centering
    \includegraphics[width=0.9\linewidth]{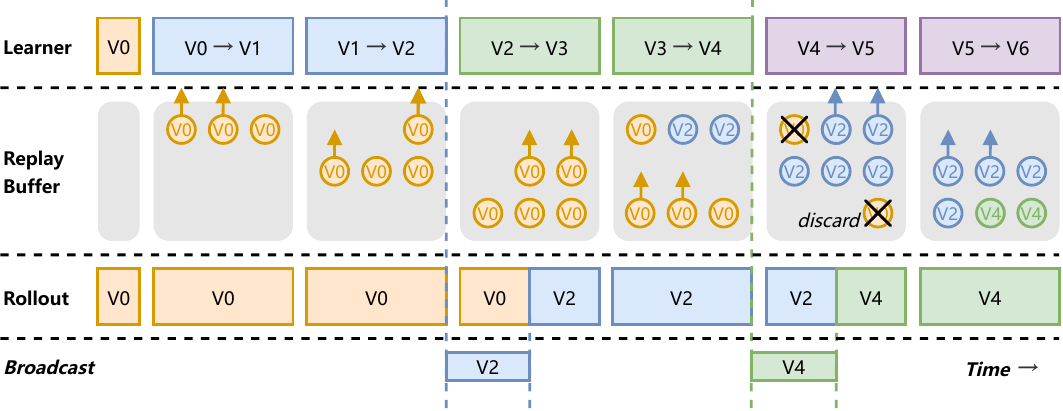}
    \caption{
    %
    Asynchronous RL execution in \sysname{}. The system overlaps rollout generation, peer-assisted policy dissemination, and learner updates. Bounded staleness $S$ and publication period $\kappa$ decouple the distributed rollout pool from the centralized learner, enabling continuous training even under wide-area network latency.
    }    
    \label{fig:async_pipeline}
\end{figure}

\section{Related Work}
\label{sec:related}

\paragraph{Asynchronous RL frameworks under controlled deployment.}
Modern RL post-training frameworks predominantly assume \emph{co-located} deployment, where learners and rollout workers share an administrative domain with provisioned high-bandwidth interconnects. \textsc{verl}~\citep{sheng2025hybridflow} supports both synchronous and asynchronous execution; AReaL~\citep{fu2025areal} streams rollouts asynchronously to hide generation variance; and AReaL-Hex~\citep{yan2025areal}, StreamRL~\citep{zhong2025streamrl}, Laminar~\citep{sheng2025laminar}, and AsyncFlow~\citep{han2025asyncflow} extend this paradigm with heterogeneous-GPU support, disaggregated stream generation, and communication optimizations. While these systems improve learner utilization under bounded asynchrony, they largely rely on data-center-grade connectivity and do not address wide-area rollout execution, where policy dissemination latency becomes a first-order cost.

\paragraph{Algorithm-level stabilization under staleness.}
A complementary line of work studies how stale rollouts alter optimization dynamics rather than system throughput. \citet{zheng2025prosperity} study how far off-policy RL can be pushed with stale data and propose M2PO, a trust-region method for large staleness; BAPO~\citep{xi2025bapo} stabilizes off-policy RL via balanced policy optimization with adaptive clipping; \citet{song2026staleness} show that stale rollouts in asynchronous GRPO induce a surrogate-gradient bias of order $O(S\eta)$ and derive a two-constraint stability rule coupling staleness $S$ and learning rate $\eta$; and GAC~\citep{xu2026gac} identifies elevated consecutive-gradient cosine similarity as a signature of asynchronous instability and introduces gradient-alignment correction under bounded staleness. These works characterize or modify the surrogate objective and optimizer-level update dynamics under a fixed system, and are orthogonal to \sysname{}: they could be layered on top of our pipeline to extend the safe range of $S$.

\paragraph{Decentralized RL.}
Closer to our setting, INTELLECT-2~\citep{team2025intellect} demonstrates fully decentralized RL by deploying both training and rollout workers across a permissionless contributor network, with health-based orchestration, Shardcast checkpoint distribution, and freshness filtering. \sysname{} instead targets a hybrid regime with centralized learning on a stable cluster and rollout inference distributed across heterogeneous wide-area workers. Its design questions are therefore different and largely orthogonal to permissionless deployment: how to expose staleness as a user-facing budget $S$, how to provision aggregate rollout capacity in closed form under dissemination overhead, and how to disseminate snapshots efficiently when the trainer uplink is the bottleneck. A detailed comparison with INTELLECT-2 is provided in \Cref{tab:intellect2-contrast} in \Cref{app:intellect2}.

\section{Cost-Efficient Distributed RL}
\label{sec:design}

\subsection{Design Rationale}
\label{sec:design:staleness}

Rollout workers differ in throughput and availability, and wide-area dissemination incurs non-negligible variable latency. Strict on-policy execution would force one side to idle, erasing the cost advantage of cheaper compute. \sysname{} instead exploits a known but underused observation: modern LLM RL objectives tolerate bounded policy lag without harming quality~\citep{zheng2025prosperity,he2025history,zhong2025streamrl,fu2025areal}. Rather than treating staleness as an artifact to minimize, we treat it as a \emph{first-class budget} that makes distributed rollouts usable.

Concretely, rollout generation, dissemination, and training proceed concurrently; the learner consumes rollouts whose policy version is at most $S$ steps stale (user-specified) and publishes a snapshot every $\kappa$ steps. \Cref{sec:overlap_condition} turns this into a provisioning question: given $(S,\kappa)$, what aggregate throughput keeps the learner saturated? The answer is a closed-form rule in the measurable quantities $(T_{\text{train}}, T_{\text{bcast}}, R, \{\mu_i\})$. Crucially, $T_{\text{bcast}}$ is itself reducible: \sysname{}'s peer-forwarding broadcast (\Cref{sec:arch:p2p}) lets workers relay snapshots and start generating immediately, so $S$ no longer merely masks network latency but controls the cost--quality trade-off.

\subsection{Execution Model and Notation}
\label{sec:notation}

The learner runs on a centralized cluster, performs one optimization step per $T_{\text{train}}$ seconds consuming $R$ trajectories, and publishes an immutable snapshot every $\kappa$ updates (larger $\kappa$ amortizes dissemination at the cost of freshness). Let $v_t$ denote the learner version at step $t$, and $v_x$ the snapshot version of the rollouts in step $t$'s batch. Define $\Delta(t)\triangleq t-x$ and $\Delta_{\max}\triangleq\max_t\Delta(t)$; the system is configured so that $\Delta_{\max}\le S$. We write $T_{\text{bcast}}$ for the learner-visible time from publication until a newly published snapshot is installed and ready to generate rollouts (measured online).

Let $\mathcal{W}$ be the set of available workers. Worker $i$ has effective throughput $\mu_i$ (rollouts/sec, end-to-end including inference, reward, scheduling, and network) and unit cost $c_i$ (\$/hour). Its \emph{unit throughput cost} is $\rho_i\triangleq c_i/\mu_i$.

\subsection{Overlap Condition and Capacity Requirement}
\label{sec:overlap_condition}

Continuous learner utilization requires that rollout generation and dissemination fit within one publication period of $\kappa$ training steps:
\begin{equation}
\kappa T_{\text{train}}
\;\ge\;
T_{\text{bcast}} + \frac{\kappa R}{\sum_{i\in\mathcal{A}} \mu_i},
\label{eq:overlap_condition}
\end{equation}
where $\mathcal{A}\subseteq\mathcal{W}$ is the active set. Rearranging gives the aggregate capacity requirement
\begin{equation}
\sum_{i\in\mathcal{A}} \mu_i
\;\ge\;
\mu_{\min}(\kappa)
\;\triangleq\;
\frac{\kappa R}{\kappa T_{\text{train}} - T_{\text{bcast}}},
\qquad
\kappa T_{\text{train}} > T_{\text{bcast}}.
\label{eq:capacity_requirement}
\end{equation}
This collapses a heterogeneous worker pool into a single measurable throughput target.

\paragraph{Linking $(S,\kappa)$.}
\Cref{app:staleness} derives a conservative upper bound assuming no rollout under a new snapshot is generated until dissemination completes:
\begin{equation*}
\Delta_{\max}^{\mathrm{cons}}
\;\le\;
\kappa
+
\big\lceil (T_{\text{bcast}}+R/\mu_{\text{pool}})/T_{\text{train}} \big\rceil
-1,
\quad
\mu_{\text{pool}}\triangleq\sum_{i\in\mathcal{A}}\mu_i.
\end{equation*}
In all our settings the overlap condition makes the ceiling term $\le 2$, so $\kappa\le S-1$ suffices and we use $\kappa = S-1$ by default. This is conservative: progressive dissemination and immediate rollout start (\Cref{sec:arch:p2p}) typically yield smaller observed $\Delta(t)$. As an example, \Cref{fig:async_pipeline} illustrates the case $S=3,\kappa=2$, in which the maximum 3-step staleness occurs at transitions $v_3\!\to\!v_4$ and $v_5\!\to\!v_6$.

\subsection{Cost-Aware Provisioning and Scheduling}
\label{sec:design:cost}

Given a candidate pool $\mathcal{W}$, \sysname{} selects $\mathcal{A}$ minimizing cost subject to the capacity rule:
\begin{equation}
\min_{\mathcal{A}\subseteq\mathcal{W}}
\sum_{i\in\mathcal{A}} c_i
\quad\text{s.t.}\quad
\sum_{i\in\mathcal{A}} \mu_i \ge \mu_{\min}(\kappa).
\label{eq:cost_provisioning}
\end{equation}
While \Cref{eq:cost_provisioning} is a knapsack-style problem, online operation favors a simple greedy: rank workers by increasing $\rho_i$ and activate the cheapest prefix whose cumulative throughput exceeds $\mu_{\min}(\kappa)$.

In practice, throughput and availability fluctuate. \sysname{} treats provisioning as closed-loop control: each worker reports lightweight statistics yielding estimates $\mu_i(t)$ and an availability bit $a_i(t)\!\in\!\{0,1\}$, with effective pool capacity
\begin{equation}
\mu_{\text{pool}}(t) \triangleq \sum_i a_i(t)\,\mu_i(t).
\label{eq:effective_pool}
\end{equation}
The scheduler targets $\mu_{\text{target}}=\gamma\mu_{\min}(\kappa)$ with $\gamma>1$ to absorb variability; if $\mu_{\text{pool}}(t)$ persistently falls below (resp.\ exceeds with margin) this target, low-$\rho$ workers are activated (resp.\ expensive ones released).

\section{System Architecture and Implementation}
\label{sec:architecture}
\begin{figure}[!t]
\centering
\includegraphics[width=0.995\linewidth]{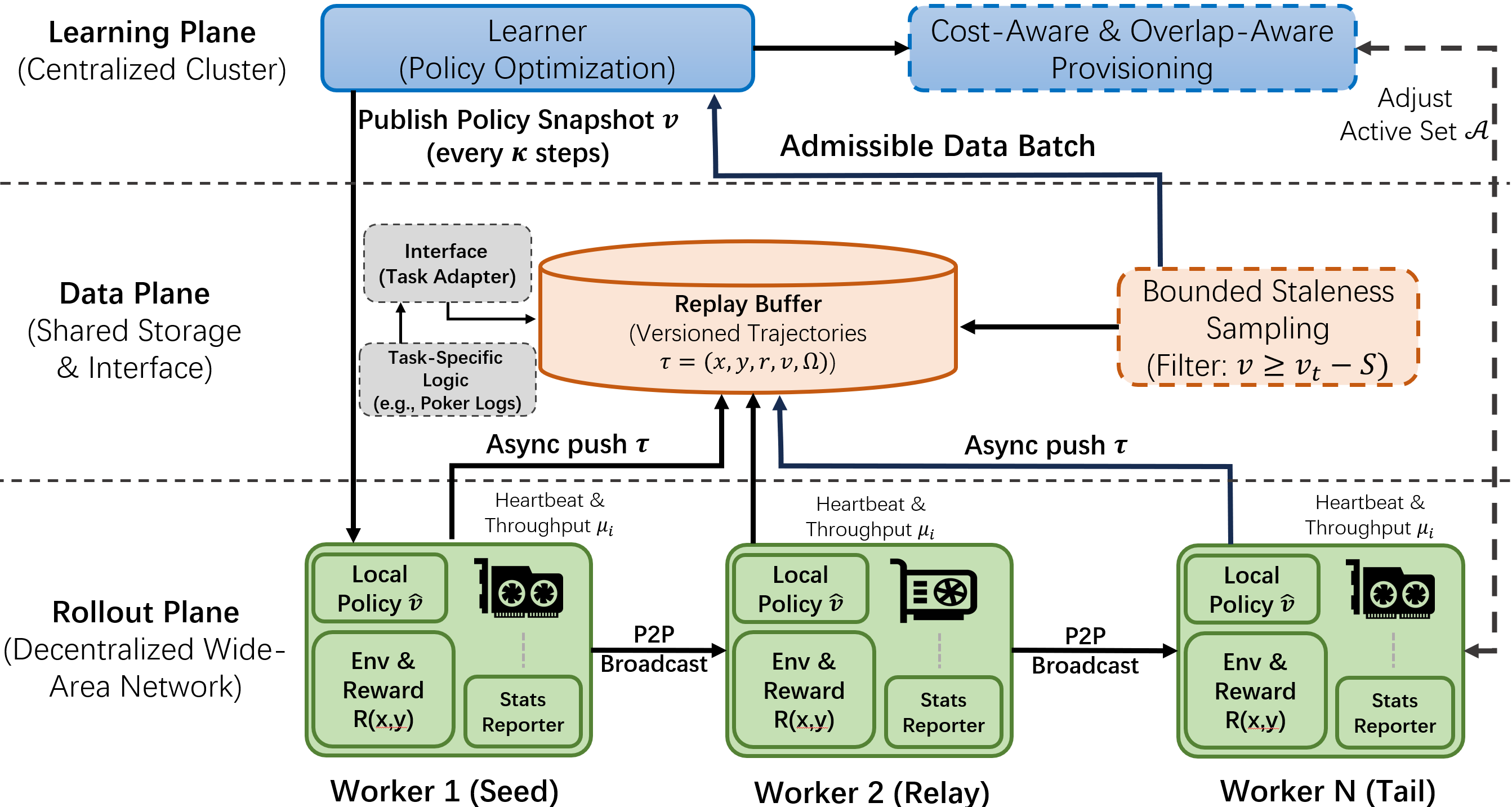}
\caption{System Architecture of \sysname{}. The system adopts a three-plane decomposition for cost-efficient distributed RL. The centralized learning plane performs policy optimization using data sampled with a bounded staleness budget. The Data Plane provides a unified interface for task adaptation and manages versioned trajectory storage. The distributed rollout plane executes asynchronous generation across workers using pipelined broadcast.
\textbf{For visual clarity, the rollout plane shows a single forwarding chain; \sysname{} runs $N_{\text{ch}} \approx \lfloor B_0 / B_w \rfloor$ such chains in parallel, with each worker maintaining one upstream parent and at most one downstream child.}}
\label{fig:arch}
\end{figure}

This section describes how \sysname{} realizes distributed and cost-efficient RL post-training with centralized learning and distributed rollouts. The design follows a three-plane decomposition: rollout, learning, and data, which are connected by versioned, immutable messages and a shared replay buffer; see Figure \ref{fig:arch}. We outline the workflow of \sysname{} in Algorithm \ref{alg:e2e}.

\textbf{Rollout Plane}: a distributed fleet of workers that repeatedly generates rewarded trajectories under a locally installed snapshot version $\hat{v}$ and pushes version-tagged results to the buffer. This plane is responsible for delivering effective throughput $\mu_i$ and forwarding policy snapshots.

\textbf{Learning Plane}: a centralized learner that consumes trajectories and performs a training step with two model updates. It enforces bounded staleness ($S$) when sampling data and publishes snapshots once every $\kappa$ learner updates. 

\textbf{Data Plane}: task adapters for prompts, trajectory schemas, reward, and loss function design. This plane provides a task-agnostic interface so new workloads are integrated by swapping datasets and reward logic, without touching scheduling or infrastructure. Detailed API specifications and a case study on poker game alignment are provided in Appendix \ref{sec:poker_case_study}.

\subsection{Versioned Execution and Bounded Staleness}
\label{sec:arch:versioned}
\paragraph{Policy publication.}
The learner publishes immutable policy snapshots once every $\kappa$ update steps (Algorithm \ref{alg:e2e} Line 15-17). Between two publications, the learner may perform multiple updates while workers continue generating rollouts under their most recent installed snapshot. The publication period $\kappa$ is chosen to respect the staleness budget $S$ (default $\kappa \le S-1$, and we use $\kappa=S-1$ unless otherwise stated), which should be $\ge2$ in \sysname{}.

\paragraph{Rollout generation.}
Each rollout worker maintains a local snapshot version $\hat{v}$. For each prompt $x$, the worker samples a response $y\sim\pi_{\hat{v}}(\cdot\mid x)$, computes reward $r=\mathcal{R}(x,y)$, and emits a trajectory $(x,y,r,\hat{v})$ into the buffer (Algorithm \ref{alg:e2e} Line 26-27). Reward computation is performed in the Rollout Plane. We reject items that violate the data format as a lightweight sanity check for data integrity.

\paragraph{Replay buffer management.}
The replay buffer stores version-tagged trajectories and supports selective sampling. At the learner update index $v_t$, only trajectories with bounded lag are admissible: $v \ge v_t - S$. Older trajectories are discarded. This enforces bounded staleness without imposing global synchronization, treating rollouts as a stream~\citep{zhong2025streamrl}.

Bounded staleness constrains data freshness but does not impose a fixed update
schedule. The learner advances whenever sufficient eligible data are available. The parameter $S$ therefore bounds the maximum policy lag between rollout generation and training, providing temporal slack to absorb latency without modifying the underlying RL objective.

\subsection{Peer-to-Peer Broadcast and Asynchronous Rollout Start}
\label{sec:arch:p2p}
A naive push-to-all strategy (star topology) achieves minimal communication latency when the bandwidth between the training center and remote workers is abundant. Under wide-area constraints, however, both the learner uplink and
the tail receivers become bottlenecks. \sysname{} instead organizes workers into a peer-forwarding topology that disseminates snapshots under bandwidth constraints by leveraging the aggregate bandwidth of the rollout fleet.

A common wide-area regime is that the learner has a finite uplink budget $B_0$, while each worker is capped by a smaller per-node bandwidth $B_w$. \sysname{} instantiates $N_{\text{ch}} \approx \lfloor B_0 / B_w \rfloor$
parallel forwarding chains, each disseminating the full snapshot via chunked store-and-forward streaming. The learner streams snapshot chunks to a first-hop seed on each chain at rate $B_w$; each seed forwards to its downstream child upon receipt, and forwarding continues along the chain. After pipeline warm-up, dissemination approaches line-rate on each chain with minimal control overhead, since each worker maintains exactly one inbound and one outbound flow ($\text{fan-out}=1$), avoiding multi-parent scheduling. Appendix \ref{app:chain-construction} details how chains are populated
and reordered online based on workers' transfer speed and reliability statistics.

To acquire as many rollouts as possible, upon receiving any new chunks, a worker forwards them immediately (Algorithm \ref{alg:e2e}, Line~22). Upon completing installation of the new snapshot, it switches its local version $\hat{v}$
and starts generating rollouts under the new version. A slow or failed relay would otherwise stall its downstream tail; \sysname{} detects such incidents through the same per-worker statistics used for chain
ranking and rebuilds the affected chain segment on the fly. Workers participating only in rollout generation are not affected by the broadcast stage faults, so a single failure causes bounded transient throughput loss rather than a pipeline stall (Appendix \ref{app:fault-tolerance}).




\subsection{Cost-Aware Scheduling over Distributed Resource Pools}
\label{sec:arch:scheduling}

\sysname{} instantiates the closed-loop scheduler of \Cref{sec:design:cost} as follows. Per-worker throughput $\mu_i(t)$ and availability $a_i(t)$ are derived from heartbeat statistics aggregated over a sliding window; $T_{\text{train}}$ and $T_{\text{bcast}}$ are calibrated offline at deployment time and re-measured only when the resource pool changes substantially. The safety factor is set to $\gamma=1.1$ in our 4B/8B experiments, giving a $\sim$10\% redundancy that absorbs typical short-term fluctuations and single-worker faults without reconfiguration (\Cref{app:fault-tolerance}). To prevent thrashing under transient noise, activation/release decisions are issued only after capacity deviates from $\mu_{\text{target}}$ for a sustained interval; the full update rule is given in \Cref{app:low_frequency_schedule}.
\section{Experiments}
\label{sec:experiments}

We evaluate \sysname{} in distributed RL post-training settings and ask three questions:
(Q1) Does \sysname{} reduce the cost required to reach a target RL quality compared with centralized pipelines?
(Q2) Is RL quality robust to bounded staleness $S$ under wide-area distributed rollouts?
(Q3) Do our overlap model and system mechanisms accurately predict and improve learner utilization under wide-area constraints?

\begin{figure*}[t]
    \centering
    \begin{subfigure}[b]{0.32\textwidth}
        \centering
        \includegraphics[width=\linewidth]{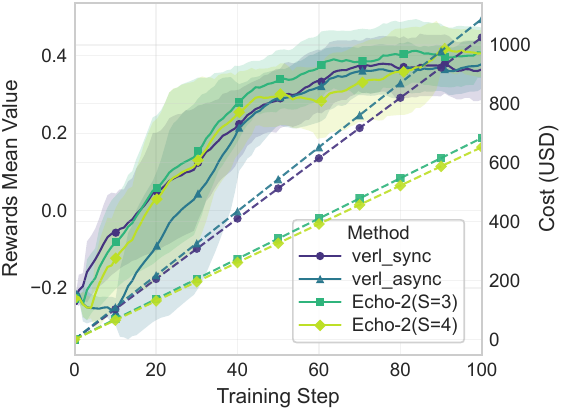}
        \caption{Cost--quality on AIME24.}
        \label{fig:costquality}
    \end{subfigure}
    \hfill
    \begin{subfigure}[b]{0.32\textwidth}
        \centering
        \includegraphics[width=\linewidth]{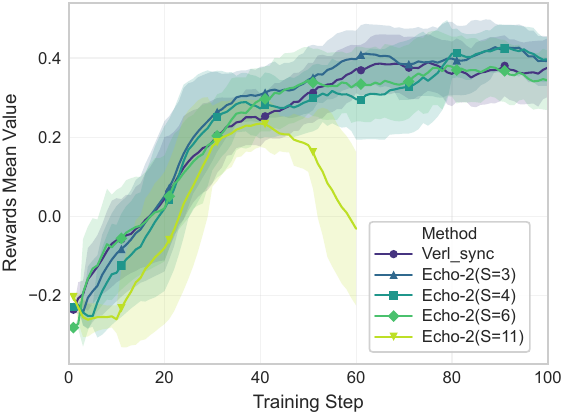}
        \caption{Effect of bounded staleness $S$.}
        \label{fig:staleness}
    \end{subfigure}
    \hfill
    \begin{subfigure}[b]{0.32\textwidth}
        \centering
        \includegraphics[width=\linewidth]{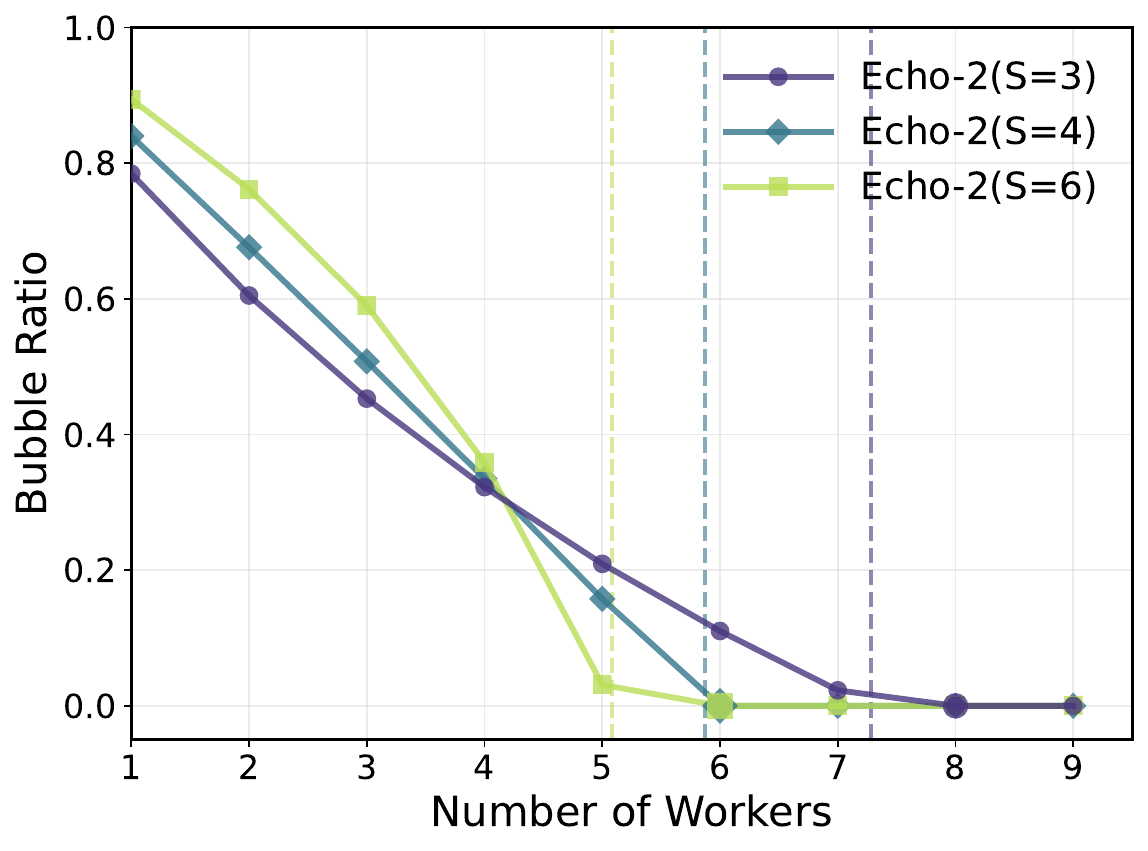}
        \caption{Bubble ratio vs.\ rollout workers.}
        \label{fig:exp:capacity}
    \end{subfigure}

    \caption{\textbf{Experimental results of \sysname{} on Qwen3-8B.}
    \label{fig:mainresults}
    \textbf{(a)} Cost-quality efficiency on AIME24 under a WAN deployment. 
    Solid lines show reward curves smoothed by a moving average over training steps; shaded regions indicate the variance of the underlying raw per-step values.
    Dashed lines indicate estimated dollar costs computed from steady-state training time and public GPU rental prices (right y-axis).
    \textbf{(b)} Effect of staleness $S$ on RL stability.
    Performance remains robust for moderate staleness ($S \le 6$), while excessive staleness ($S=11$) leads to divergence.
    \textbf{(c)} Learner bubble ratio as a function of the number of rollout workers.
    Vertical dashed lines denote the theoretically predicted minimum worker counts.}
    \label{fig:main_results}
\end{figure*}

\subsection{Experimental Setup}
\label{sec:exp:setup}

\textbf{Models.}
We post-train two base models, Qwen3-4B and Qwen3-8B~\citep{qwen3technicalreport}.
Unless otherwise stated, we use the same GRPO hyperparameters across all systems:
global batch size $128$, maximum generation length $8192$, temperature $1.0$, top-$p$ $0.95$, and rollout multiplicity $n=16$.
We disable chain-of-thought prompting and report avg@64.

\textbf{Datasets.}
We train on DAPO-Math~\citep{yu2025dapoopensourcellmreinforcement} and evaluate on AIME24~\citep{maa2024aime}, which has verifiable final answers.
Unless otherwise noted, we report AIME accuracy as the primary RL quality metric.
We additionally report results on a broader math benchmark suite in Appendix \ref{app:extra_bench}.

\textbf{System deployment.}
(1) \emph{Learning plane.}
The learner runs on $4\times$ A100 80GB GPUs for \sysname{}, and on $8\times$ A100 GPUs for the centralized baseline.
We measure the steady-state per-update time $T_{\text{train}}$ as the median over five offline training steps.
(2) \emph{Rollout plane.}
To simplify experimental validation, rollouts run on a distributed pool of RTX 5090 workers served by Parallax~\citep{tong2025parallax}, a hardware-agnostic inference service.

\textbf{Network regimes.}
To study the impact of bandwidth constraints, we cap the learner uplink budget to representative WAN regimes and cap each worker's download rate.
In experiments, we identify $T_{\text{bcast}}$ as the elapsed time from publication until a target fraction $1/\gamma$ of active workers have installed the snapshot.
This aligns the abstraction in \Cref{sec:design} with the observed dissemination behavior and captures the practical effect of tail receivers.

\textbf{Baselines.}
(1) \textsc{Centralized-Sync} (\textsc{veRL}~\citep{sheng2025hybridflow}): a synchronized pipeline in which rollouts are co-located with the learner on the same data-center GPUs.
(2) \textsc{Centralized-Async} (\textsc{veRL}-async~\citep{sheng2025hybridflow}): a streaming asynchronous baseline within the data center (AReaL-style~\citep{fu2025areal}) that overlaps rollout generation and learning while assuming high-bandwidth, low-latency connectivity.



\begin{wraptable}{r}{0.45\textwidth}
\vspace{-1em}
\centering
\small
\caption{GPU rental prices (USD/hour, single GPU, collected on 2026-05-01).}
\label{tab:cost_sources}
\begin{tabular}{lcc}
\toprule
\textbf{GPU} & \textbf{USD/hr} & \textbf{Platform} \\
\midrule
A100 80GB    & \$3.06          & \href{https://gcloud-compute.com/a2-ultragpu-1g.html}{Google Cloud} \\
RTX 5090     & \$0.35          & \href{https://vast.ai/pricing}{vast.ai} \\
\bottomrule
\end{tabular}
\vspace{-1em}
\end{wraptable}


\textbf{Cost and utilization metrics.}
Dollar cost is the sum over GPU types of hourly rental price (\Cref{tab:cost_sources}) times GPU-hours used, with Google Cloud A100 representing data-center hardware and vast.ai RTX~5090 approximating distributed customer-grade resources. The learner \emph{bubble ratio} is the fraction of wall-clock time the learner spends idle waiting for admissible rollouts.
We measure the learner \emph{bubble ratio} (idle fraction) as
\[
\frac{T_{\text{idle}}}{T_{\text{idle}} + T_{\text{train,active}}},
\]
where $T_{\text{idle}}$ is the waiting time caused by insufficient admissible rollouts.

\subsection{Cost--Quality Efficiency}
\label{sec:exp:costquality}

Given the discreteness and variance of AIME accuracy, we interpret cost--quality efficiency as the cumulative cost required to reach a target accuracy threshold under a representative WAN setting. \Cref{fig:costquality} plots the reward curves; the per-step training times ($T_{\text{train}}$) of \textsc{veRL}-sync, \textsc{veRL}-async, and \sysname{} with $S=3/4$ are 1508.2s, 1582.3s, 1631.2s, and 1649.3s, respectively. Combining $T_{\text{train}}$ with rental prices under steady-state execution yields the linear cost curves shown as dashed lines (right $y$-axis).

For Qwen3-8B, \sysname{} consistently dominates centralized pipelines: at matched AIME accuracy, it reduces cumulative cost by 33.3\%--36.3\%; at matched cost, it achieves comparable final accuracy. This gain arises because rollout generation is offloaded to cheaper distributed GPUs without stalling the centralized learner, as long as the overlap condition (\Cref{eq:overlap_condition}) is satisfied. Qwen3-4B shows the same trend and leads to the same conclusion (\Cref{app:result_4b}).

\paragraph{Generalization to larger models.}
We additionally validate \sysname{} on QwQ-32B against Prime-RL~\citep{team2025intellect}, a recent decentralized-rollout system, under matched GRPO hyperparameters. Despite running rollouts on $76\times$RTX~5090 workers over a wide-area network (vs.\ Prime-RL's $10\times$H100 on an idealized intranet), \sysname{} achieves a 4.6\% end-to-end cost reduction with reward trajectories tracking Prime-RL throughout training. This isolates the contribution of our system mechanisms (overlap-aware provisioning, peer-forwarding broadcast, cost-aware activation) from the architectural shift to disaggregated rollout that Prime-RL also adopts. Notably, this saving is achieved \emph{despite} \sysname{} running over a wide-area network while Prime-RL enjoys an idealized intranet with unlimited trainer-to-worker bandwidth; under a fair WAN comparison, Prime-RL's centralized broadcast cost would scale with fleet size while \sysname{}'s remains bounded (\Cref{sec:exp:broadcast}). Full setup and stability analysis are in \Cref{app:32b-prime-rl}.

\subsection{RL Quality under Bounded Staleness}
\label{sec:exp:staleness}

We sweep the staleness budget $S$ to quantify the trade-off between training stability, final quality, and system efficiency.
Specifically, we evaluate $S \in \{3,4,6,11\}$ while keeping all other settings fixed.
Recall that, in a stable pipelined \sysname{}, the observed rollout staleness is upper-bounded by $S$.

\Cref{fig:staleness} shows that moderate staleness does not materially degrade final quality:
for $S \le 6$, \sysname{} achieves reward and accuracy within roughly 5\% fluctuation of the synchronous baseline, with similar convergence trends and lower cost.
In contrast, overly large staleness ($S=11$) can lead to instability in standard GRPO, consistent with the intuition that stale data gradually deviates from the current policy distribution, and with the analysis of \citet{song2026staleness}, which shows that the maximum stable learning rate shrinks roughly as $1/S$.

\subsection{Validating the Overlap Condition}
\label{sec:exp:capacity}

\Cref{eq:capacity_requirement} predicts a threshold behavior:
as rollout capacity increases, learner bubbles should rapidly vanish once the system enters the feasible overlap region.
We validate this prediction by sweeping the effective rollout-pool size.
\Cref{fig:exp:capacity} shows that, as the pool grows, bubble ratio drops consistently toward zero near the predicted threshold, confirming that the overlap model provides a practical provisioning rule.
Larger $S$ shifts the transition to the left, showing that staleness acts as an explicit control knob that trades policy freshness for reduced rollout capacity.

\begin{figure}[t]
\centering
\includegraphics[width=0.8\linewidth]{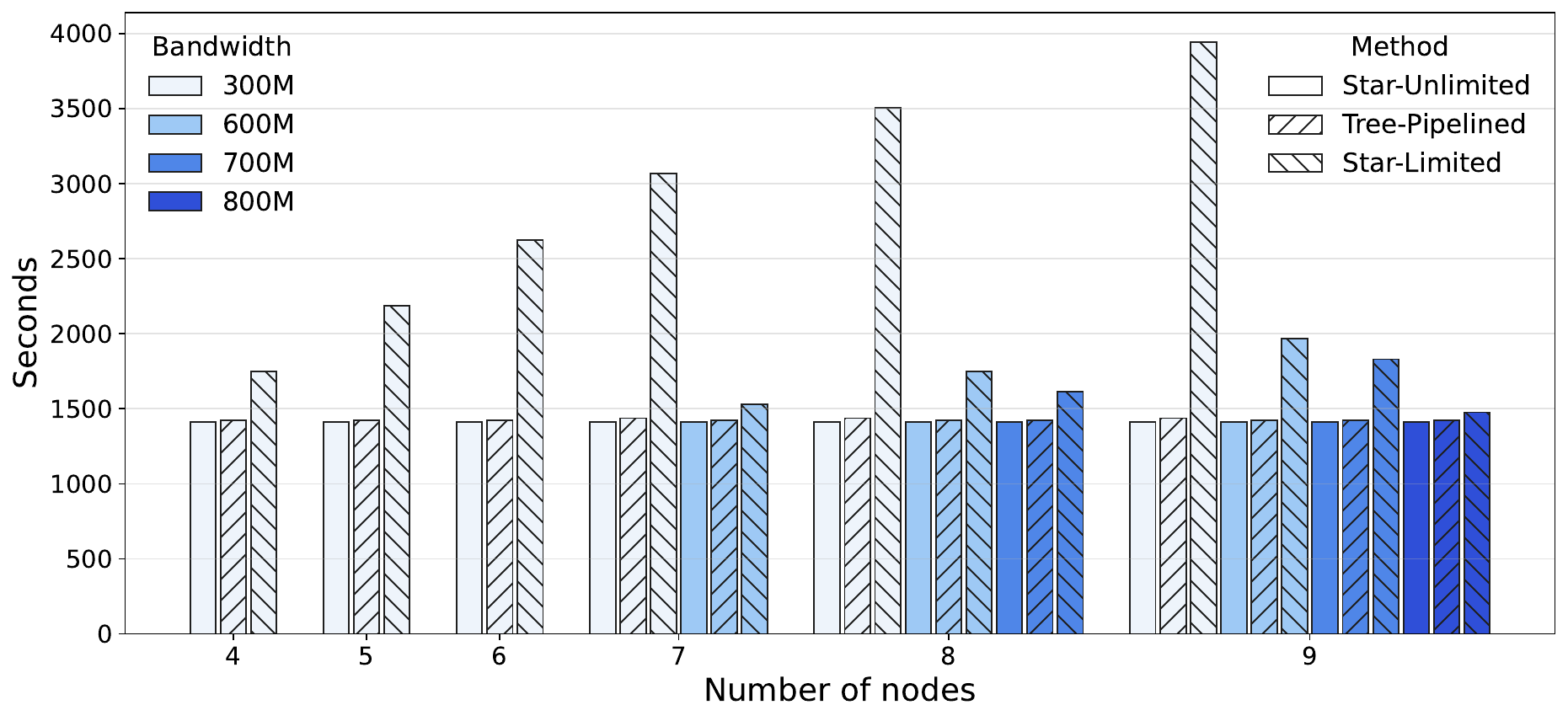}
\caption{\textbf{Policy broadcast latency $T_{\text{bcast}}$ vs.\ rollout fleet size.}
Comparison of three dissemination strategies across different numbers of nodes $N$.
\textbf{Star-Limited} (with learner uplink $B_0 \in [300, 800]$ Mbps) suffers from increasing latency as the learner becomes the bandwidth bottleneck.
\textbf{Tree-Pipelined} dissemination, by using chunked peer forwarding, maintains near-constant broadcast time and scales much more efficiently with fleet size, closely matching the idealized \textbf{Star-Unlimited} baseline.}
\label{fig:exp:broadcast}
\end{figure}

\subsection{Ablation Study}
\label{sec:exp:ablation}

\subsubsection{Broadcast under Bandwidth Constraints}
\label{sec:exp:broadcast}

We evaluate policy dissemination latency under different broadcast strategies
as the rollout fleet scales. We measure the learner-visible broadcast time
$T_{\text{bcast}}$ as the elapsed time from snapshot publication until a target
fraction $q$ of active workers have fully installed the snapshot and can start
generating rollouts under the new version (\Cref{sec:arch:p2p}). We use
$q = 1/\gamma = 1/1.1 \approx 0.9$.

We compare three dissemination settings while sweeping the number of active
rollout workers $N$:
(i) \textsc{Star-Unlimited}, an idealized push-to-all broadcast with no learner
uplink cap, serving as a lower bound on $T_{\text{bcast}}$;
(ii) \textsc{Star-Limited}, push-to-all broadcast under realistic wide-area
caps, with the learner uplink budget $B_0$ swept over $\{300, 600, 700, 800\}$
Mbps and a per-worker downlink cap $B_w = 100$ Mbps; and
(iii) \textsc{Tree-Pipelined}, ECHO-2's design, which constructs $N_{\text{ch}}
\approx \lfloor B_0 / B_w \rfloor$ parallel forwarding chains and uses chunked
store-and-forward dissemination so that each worker relays data to its single
downstream child upon receipt (\Cref{sec:arch:p2p}).

\Cref{fig:exp:broadcast} shows that \textsc{Star-Limited} suffers from rapidly
growing $T_{\text{bcast}}$ as $N$ increases: with a fixed learner uplink budget
$B_0$, the per-worker delivery rate degrades to $B_0 / N$ once $N \cdot B_w >
B_0$, making the learner the dominant bottleneck. In contrast,
\textsc{Tree-Pipelined} keeps $T_{\text{bcast}}$ within a small constant factor
of \textsc{Star-Unlimited} across the entire range, even under the same
bandwidth caps. This is because each chain operates at line rate $B_w$
independently after pipeline warm-up, and the aggregate fleet bandwidth scales
with $N$ rather than being capped by $B_0$.

\subsubsection{Peer-Assisted Broadcast and Cost-Aware Provisioning}



\begin{wraptable}{r}{0.5\textwidth}

\centering
\small
\caption{\textbf{Ablation summary.} We evaluate the impact of removing peer-assisted (P2P) broadcast and cost-aware provisioning (Cost) in \sysname{}. \#Mach, $T_{\text{bcast}}$, and Wait denote the rollout fleet size, dissemination latency, and learner idle time, respectively.}

\label{tab:ablation}
\setlength{\tabcolsep}{4pt} 
\begin{tabular}{lcccc}
\toprule
Method & \#Mach & Cost/Step $\downarrow$ & $T_{\text{bcast}}$ (s) & Wait (s) \\
\midrule
Full & 9 & 8.098 & 1437 & 0 \\
\midrule
\textit{w/o} P2P & 9 & 8.432 & 1830 & 131.9 \\
\textit{w/o} P2P & 10 & 8.630 & 1872 & 84.1 \\
\textit{w/o} Cost & 9 & 9.339 & 1437 & 0 \\
\bottomrule
\end{tabular}

\end{wraptable}

We isolate the contributions of \sysname{}'s two mechanisms via two ablations: \sysname-\textsc{NoP2P} replaces peer-assisted broadcast with direct learner-to-worker dissemination, and \sysname-\textsc{NoCost} replaces cost-aware provisioning with uniform random activation while still targeting the full-overlap objective. To make the cost ablation informative, we evaluate under a mixed-price rollout pool with heterogeneous worker costs, and report cost per step, dissemination latency, and inter-step learner waiting time.

\Cref{tab:ablation} shows that removing peer-assisted broadcast inflates communication latency and bubble time, requiring additional machines---and thus higher cost---to suppress learner idle time. Disabling cost-aware provisioning likewise increases the cost to reach a given quality target by activating suboptimal workers. Together, these results show that both mechanisms are necessary for the end-to-end cost efficiency reported in \Cref{fig:costquality}. A task-agnostic data-plane use case is discussed in \Cref{sec:poker_case_study}.






\section{Limitations and Future Work}
\label{sec:limitations}

\sysname{} relies on the empirical robustness of modern LLM RL objectives to bounded policy lag. While moderate staleness preserves GRPO post-training quality in our experiments, we do not provide formal guarantees, and the safe range may depend on the task and reward signal. Developing theoretical staleness control remains future work. Our peer-assisted broadcast mitigates uplink bottlenecks, whereas our future work will include delta or quantized updates and cache-aware deployment. \sysname{} focuses on centralized learning with distributed rollouts.
Extending the design to multiple or geographically replicated learners is promising, but it introduces new challenges in synchronization and policy consistency, and further validation across a wider range of model sizes (70B+) remains future work.

\section{Conclusion}
\label{sec:conclusion} 
We presented \sysname{}, an RL framework for LLM post-training that separates centralized learning from distributed rollouts. By treating bounded staleness as a control knob, modeling overlap-based capacity, and activating workers on demand based on cost, \sysname{} enables cost-aware provisioning under wide-area execution. With peer-assisted pipelined broadcast, \sysname{} reduces dissemination overhead. Experiments on GRPO post-training of 4B, 8B and 32B models show that \sysname{} significantly lowers training cost while preserving RL quality comparable to baselines.

\bibliographystyle{plainnat}
\bibliography{refs}

@inproceedings{sheng2025hybridflow,
  title={Hybridflow: A flexible and efficient rlhf framework},
  author={Sheng, Guangming and Zhang, Chi and Ye, Zilingfeng and Wu, Xibin and Zhang, Wang and Zhang, Ru and Peng, Yanghua and Lin, Haibin and Wu, Chuan},
  booktitle={Proceedings of the Twentieth European Conference on Computer Systems},
  pages={1279--1297},
  year={2025}
}

@article{xiao2025echo,
  title={Echo: Decoupling Inference and Training for Large-Scale RL Alignment on Heterogeneous Swarms},
  author={Xiao, Jie and Fan, Changyuan and Ren, Qingnan and Long, Alfred and Zhang, Yuchen and Yu, Rymon and Yang, Eric and Ai, Lynn and Gan, Shaoduo},
  journal={arXiv preprint arXiv:2508.05387},
  year={2025}
}

@inproceedings{borzunov2023petals,
  title={Petals: Collaborative inference and fine-tuning of large models},
  author={Borzunov, Alexander and Baranchuk, Dmitry and Dettmers, Tim and Riabinin, Maksim and Belkada, Younes and Chumachenko, Artem and Samygin, Pavel and Raffel, Colin},
  booktitle={Proceedings of the 61st Annual Meeting of the Association for Computational Linguistics (Volume 3: System Demonstrations)},
  pages={558--568},
  year={2023}
}

@inproceedings{ryabinin2023swarm,
  title={Swarm parallelism: Training large models can be surprisingly communication-efficient},
  author={Ryabinin, Max and Dettmers, Tim and Diskin, Michael and Borzunov, Alexander},
  booktitle={International Conference on Machine Learning},
  pages={29416--29440},
  year={2023},
  organization={PMLR}
}

@misc{schulman2017proximalpolicyoptimizationalgorithms,
      title={Proximal Policy Optimization Algorithms}, 
      author={John Schulman and Filip Wolski and Prafulla Dhariwal and Alec Radford and Oleg Klimov},
      year={2017},
      eprint={1707.06347},
      archivePrefix={arXiv},
      primaryClass={cs.LG},
      url={https://arxiv.org/abs/1707.06347}, 
}

@article{fu2025areal,
  title={AReaL: A Large-Scale Asynchronous Reinforcement Learning System for Language Reasoning},
  author={Fu, Wei and Gao, Jiaxuan and Shen, Xujie and Zhu, Chen and Mei, Zhiyu and He, Chuyi and Xu, Shusheng and Wei, Guo and Mei, Jun and Wang, Jiashu and others},
  journal={arXiv preprint arXiv:2505.24298},
  year={2025}
}

@article{yan2025areal,
  title={AReaL-Hex: Accommodating Asynchronous RL Training over Heterogeneous GPUs},
  author={Yan, Ran and Jiang, Youhe and Wu, Tianyuan and Gao, Jiaxuan and Mei, Zhiyu and Fu, Wei and Mai, Haohui and Wang, Wei and Wu, Yi and Yuan, Binhang},
  journal={arXiv preprint arXiv:2511.00796},
  year={2025}
}

@article{team2025intellect,
  title={INTELLECT-2: A Reasoning Model Trained Through Globally Decentralized Reinforcement Learning},
  author={Team, Prime Intellect and Jaghouar, Sami and Mattern, Justus and Ong, Jack Min and Straube, Jannik and Basra, Manveer and Pazdera, Aaron and Thaman, Kushal and Di Ferrante, Matthew and Gabriel, Felix and others},
  journal={arXiv preprint arXiv:2505.07291},
  year={2025}
}

@article{noukhovitch2024asynchronous,
  title={Asynchronous rlhf: Faster and more efficient off-policy rl for language models},
  author={Noukhovitch, Michael and Huang, Shengyi and Xhonneux, Sophie and Hosseini, Arian and Agarwal, Rishabh and Courville, Aaron},
  journal={arXiv preprint arXiv:2410.18252},
  year={2024}
}

@article{team2025kimi,
  title={Kimi k1. 5: Scaling reinforcement learning with llms},
  author={Team, Kimi and Du, Angang and Gao, Bofei and Xing, Bowei and Jiang, Changjiu and Chen, Cheng and Li, Cheng and Xiao, Chenjun and Du, Chenzhuang and Liao, Chonghua and others},
  journal={arXiv preprint arXiv:2501.12599},
  year={2025}
}

@article{guo2025deepseek,
  title={Deepseek-r1: Incentivizing reasoning capability in llms via reinforcement learning},
  author={Guo, Daya and Yang, Dejian and Zhang, Haowei and Song, Junxiao and Zhang, Ruoyu and Xu, Runxin and Zhu, Qihao and Ma, Shirong and Wang, Peiyi and Bi, Xiao and others},
  journal={arXiv preprint arXiv:2501.12948},
  year={2025}
}

@article{shao2024deepseekmath,
  title={Deepseekmath: Pushing the limits of mathematical reasoning in open language models},
  author={Shao, Zhihong and Wang, Peiyi and Zhu, Qihao and Xu, Runxin and Song, Junxiao and Bi, Xiao and Zhang, Haowei and Zhang, Mingchuan and Li, YK and Wu, Yang and others},
  journal={arXiv preprint arXiv:2402.03300},
  year={2024}
}

@article{he2025history,
  title={History rhymes: Accelerating llm reinforcement learning with rhymerl},
  author={He, Jingkai and Li, Tianjian and Feng, Erhu and Du, Dong and Liu, Qian and Liu, Tao and Xia, Yubin and Chen, Haibo},
  journal={arXiv preprint arXiv:2508.18588},
  year={2025}
}

@article{zhong2025streamrl,
  title={StreamRL: Scalable, Heterogeneous, and Elastic RL for LLMs with Disaggregated Stream Generation},
  author={Zhong, Yinmin and Zhang, Zili and Song, Xiaoniu and Hu, Hanpeng and Jin, Chao and Wu, Bingyang and Chen, Nuo and Chen, Yukun and Zhou, Yu and Wan, Changyi and others},
  journal={arXiv preprint arXiv:2504.15930},
  year={2025}
}

@article{tong2025parallax,
  title={Parallax: Efficient LLM Inference Service over Decentralized Environment},
  author={Tong, Chris and Jiang, Youhe and Chen, Gufeng and Zhao, Tianyi and Lu, Sibian and Qu, Wenjie and Yang, Eric and Ai, Lynn and Yuan, Binhang},
  journal={arXiv preprint arXiv:2509.26182},
  year={2025}
}

@article{zhou2025rlax,
  title={RLAX: Large-Scale, Distributed Reinforcement Learning for Large Language Models on TPUs},
  author={Zhou, Runlong and Zhang, Lefan and Wu, Shang-Chen and Zou, Kelvin and Zhou, Hanzhi and Ye, Ke and Feng, Yihao and Yin, Dong and Garcia, Alex Guillen and Babych, Dmytro and others},
  journal={arXiv preprint arXiv:2512.06392},
  year={2025}
}

@article{zheng2025prosperity,
  title={Prosperity before Collapse: How Far Can Off-Policy RL Reach with Stale Data on LLMs?},
  author={Zheng, Haizhong and Zhao, Jiawei and Chen, Beidi},
  journal={arXiv preprint arXiv:2510.01161},
  year={2025}
}

@misc{qwen3technicalreport,
      title={Qwen3 Technical Report}, 
      author={Qwen Team},
      year={2025},
      eprint={2505.09388},
      archivePrefix={arXiv},
      primaryClass={cs.CL},
      url={https://arxiv.org/abs/2505.09388}, 
}

@misc{comanici2025gemini25pushingfrontier,
      title={Gemini 2.5: Pushing the Frontier with Advanced Reasoning, Multimodality, Long Context, and Next Generation Agentic Capabilities}, 
      author={Gheorghe Comanici and Eric Bieber and Mike Schaekermann and others},
      year={2025},
      eprint={2507.06261},
      archivePrefix={arXiv},
      primaryClass={cs.CL},
      url={https://arxiv.org/abs/2507.06261}, 
}

@misc{openai2025gpt5,
  title        = {OpenAI GPT-5 System Card},
  author       = {{OpenAI}},
  year         = {2025},
  eprint       = {2601.03267},
  archivePrefix= {arXiv},
  primaryClass = {cs.CL},
  url          = {https://arxiv.org/abs/2601.03267}
}

@techreport{xai2025grok4,
  title        = {Grok 4 Model Card},
  author       = {{xAI}},
  institution  = {xAI},
  year         = {2025},
  url          = {https://data.x.ai/2025-08-20-grok-4-model-card.pdf}
}

@techreport{anthropic2025claude45,
  title        = {Claude Sonnet 4.5 System Card},
  author       = {{Anthropic}},
  institution  = {Anthropic},
  year         = {2025},
  url          = {https://assets.anthropic.com/m/12f214efcc2f457a/original/Claude-Sonnet-4-5-System-Card.pdf}
}

@misc{maa2024aime, 
  title={American Invitational Mathematics Examination ({AIME})},
  author={{MAA}},
  year={2024},
  url={https://www.maa.org/math-competitions/aime}
}

@article{gao2024omni,
  title={Omni-math: A universal olympiad level mathematic benchmark for large language models},
  author={Gao, Bofei and Song, Feifan and Yang, Zhe and Cai, Zefan and Miao, Yibo and Dong, Qingxiu and Li, Lei and Ma, Chenghao and Chen, Liang and Xu, Runxin and others},
  journal={arXiv preprint arXiv:2410.07985},
  year={2024}
}

@inproceedings{luong2025towards,
  title={Towards robust mathematical reasoning},
  author={Luong, Minh-Thang and Hwang, Dawsen and Nguyen, Hoang H and Ghiasi, Golnaz and Chervonyi, Yuri and Seo, Insuk and Kim, Junsu and Bingham, Garrett and Lee, Jonathan and Mishra, Swaroop and others},
  booktitle={Proceedings of the 2025 Conference on Empirical Methods in Natural Language Processing},
  pages={35406--35430},
  year={2025}
}

@article{fan2024hardmath,
  title={Hardmath: A benchmark dataset for challenging problems in applied mathematics},
  author={Fan, Jingxuan and Martinson, Sarah and Wang, Erik Y and Hausknecht, Kaylie and Brenner, Jonah and Liu, Danxian and Peng, Nianli and Wang, Corey and Brenner, Michael P},
  journal={arXiv preprint arXiv:2410.09988},
  year={2024}
}

@inproceedings{arora2023have,
  title={Have llms advanced enough? a challenging problem solving benchmark for large language models},
  author={Arora, Daman and Singh, Himanshu and others},
  booktitle={Proceedings of the 2023 Conference on Empirical Methods in Natural Language Processing},
  pages={7527--7543},
  year={2023}
}

@misc{yu2025dapoopensourcellmreinforcement,
      title={DAPO: An Open-Source LLM Reinforcement Learning System at Scale}, 
      author={Qiying Yu and Zheng Zhang and Ruofei Zhu and Yufeng Yuan and Xiaochen Zuo and Yu Yue and Tiantian Fan and Gaohong Liu and Lingjun Liu and Xin Liu and Haibin Lin and Zhiqi Lin and Bole Ma and Guangming Sheng and Yuxuan Tong and Chi Zhang and Mofan Zhang and Wang Zhang and Hang Zhu and Jinhua Zhu and Jiaze Chen and Jiangjie Chen and Chengyi Wang and Hongli Yu and Weinan Dai and Yuxuan Song and Xiangpeng Wei and Hao Zhou and Jingjing Liu and Wei-Ying Ma and Ya-Qin Zhang and Lin Yan and Mu Qiao and Yonghui Wu and Mingxuan Wang},
      year={2025},
      eprint={2503.14476},
      archivePrefix={arXiv},
      primaryClass={cs.LG},
      url={https://arxiv.org/abs/2503.14476}, 
}

@misc{xu2026gac,
      title={GAC: Stabilizing Asynchronous RL Training for LLMs via Gradient Alignment Control}, 
      author={Haofeng Xu and Junwei Su and Yukun Tian and Lansong Diao and Zhengping Qian and Chuan Wu},
      year={2026},
      eprint={2603.01501},
      archivePrefix={arXiv},
      primaryClass={cs.LG},
      url={https://arxiv.org/abs/2603.01501}, 
}

@misc{han2025asyncflow,
      title={AsyncFlow: An Asynchronous Streaming RL Framework for Efficient LLM Post-Training}, 
      author={Zhenyu Han and Ansheng You and Haibo Wang and Kui Luo and Guang Yang and Wenqi Shi and Menglong Chen and Sicheng Zhang and Zeshun Lan and Chunshi Deng and Huazhong Ji and Wenjie Liu and Yu Huang and Yixiang Zhang and Chenyi Pan and Jing Wang and Xin Huang and Chunsheng Li and Jianping Wu},
      year={2025},
      eprint={2507.01663},
      archivePrefix={arXiv},
      primaryClass={cs.LG},
      url={https://arxiv.org/abs/2507.01663}, 
}

@misc{sheng2025laminar,
      title={Laminar: A Scalable Asynchronous RL Post-Training Framework}, 
      author={Guangming Sheng and Yuxuan Tong and Borui Wan and Wang Zhang and Chaobo Jia and Xibin Wu and Yuqi Wu and Xiang Li and Chi Zhang and Yanghua Peng and Haibin Lin and Xin Liu and Chuan Wu},
      year={2025},
      eprint={2510.12633},
      archivePrefix={arXiv},
      primaryClass={cs.LG},
      url={https://arxiv.org/abs/2510.12633}, 
}

@misc{xi2025bapo,
      title={BAPO: Stabilizing Off-Policy Reinforcement Learning for LLMs via Balanced Policy Optimization with Adaptive Clipping}, 
      author={Zhiheng Xi and Xin Guo and Yang Nan and Enyu Zhou and Junrui Shen and Wenxiang Chen and Jiaqi Liu and Jixuan Huang and Zhihao Zhang and Honglin Guo and Xun Deng and Zhikai Lei and Miao Zheng and Guoteng Wang and Shuo Zhang and Peng Sun and Rui Zheng and Hang Yan and Tao Gui and Qi Zhang and Xuanjing Huang},
      year={2025},
      eprint={2510.18927},
      archivePrefix={arXiv},
      primaryClass={cs.LG},
      url={https://arxiv.org/abs/2510.18927}, 
}

@misc{song2026staleness,
      title={Scaling Laws for Collapse in Asynchronous GRPO}, 
      author={Jingwei Song and Haofeng Xu and Jie Xiao and Chengke Bao and Jingwei Shi and Pengbin Feng and Yuhang Han and Weixun Wang and Eric Yang and Tianyu Shi},
      year={2026},
      eprint={2607.01083},
      archivePrefix={arXiv},
      primaryClass={cs.LG},
      url={https://arxiv.org/abs/2607.01083}, 
}


\appendix




\newpage

\section{Detailed Comparison with INTELLECT-2}
\label{app:intellect2}

\Cref{tab:intellect2-contrast} summarizes the design contrast between INTELLECT-2 and \sysname{} along four dimensions. The two systems target orthogonal sets of design questions: INTELLECT-2 focuses on infrastructure for permissionless contributor participation, while \sysname{} focuses on cost-efficient hybrid execution under bounded staleness with a closed-form provisioning rule.

\begin{table}[h]
\centering
\small
\caption{Design contrast between INTELLECT-2 and \sysname{}.}
\label{tab:intellect2-contrast}
\begin{tabular}{lll}
\toprule
\textbf{Design dimension} & \textbf{INTELLECT-2} & \textbf{\sysname{} (ours)} \\
\midrule
Policy staleness   & Empirical at fixed asynchrony            & User-facing budget $S$; $\kappa$ from $S$ via \Cref{app:staleness} \\
Provisioning rule  & Case-by-case fit check                   & Closed-form $\sum\mu_i \ge \mu_{\min}(\kappa)$ \\
Worker activation  & Health/availability-based                & Cost-aware via $\rho_i = c_i/\mu_i$ \\
Broadcast topology & Shardcast (CDN-style relay tier)         & Peer-forwarding chains within fleet \\
\bottomrule
\end{tabular}
\end{table}

\section{Worst-Case Staleness Bound under Overlap}
\label{app:staleness}

This appendix derives a conservative upper bound on the maximum policy staleness
$\Delta_{\max}$ in \sysname{}, and shows how the overlap condition tightens this
bound.

\subsection{Execution Semantics and Conservative Assumption}

We consider the following execution semantics, which intentionally model a
worst-case scenario:

\begin{itemize}
    \item Policy snapshots are published every $\kappa$ learner update steps, at
    the \emph{end} of a training step.
    \item Training batches are formed at the \emph{beginning} of each step.
    \item In the most conservative case, rollout workers generate no trajectories
    from a newly published snapshot until dissemination completes after
    $T_{\text{bcast}}$ time.
    \item After dissemination completes, rollouts from the new policy are
    generated at aggregated rate $\mu_{\text{pool}}$.
\end{itemize}

This model intentionally ignores progressive dissemination and early rollout
start, and therefore upper-bounds the staleness that can occur in practice.

\subsection{Baseline Worst-Case Staleness Bound}

Let $n$ denote the number of learner steps elapsed since a snapshot is published.
By time $nT_{\text{train}}$, the number of rollouts generated from the new policy
is at most
\begin{equation}
G(n)
=
\mu_{\text{pool}}
\cdot
\max\!\bigl(0,\; nT_{\text{train}} - T_{\text{bcast}}\bigr).
\end{equation}

The earliest step at which at least $R$ new-policy rollouts are available satisfies
\begin{equation}
G(n) \;\ge\; R
\quad\Rightarrow\quad
n \;\ge\;
\left\lceil
\frac{T_{\text{bcast}} + \frac{R}{\mu_{\text{pool}}}}{T_{\text{train}}}
\right\rceil.
\end{equation}

At publication, the learner version advances by $\kappa$ relative to the previous
published snapshot.
Since no new-policy rollout can be consumed during the first $n-1$ steps after
publication, the maximum staleness under this conservative model is
\begin{equation}
\Delta_{\max}^{\mathrm{cons}}
=
\kappa
+
\left\lceil
\frac{T_{\text{bcast}} + \frac{R}{\mu_{\text{pool}}}}{T_{\text{train}}}
\right\rceil
-1.
\label{eq:delta_cons}
\end{equation}

\subsection{Tightening the Bound using the Overlap Condition}

The baseline bound in \Cref{eq:delta_cons} depends on the rollout throughput
$\mu_{\text{pool}}$.
We now show that under the overlap condition, this dependence can be tightened.

Recall the overlap condition:
\begin{equation}
\kappa T_{\text{train}}
\;\ge\;
T_{\text{bcast}}
+
\frac{\kappa R}{\mu_{\text{pool}}}.
\label{eq:overlap_app}
\end{equation}

Rearranging yields
\begin{equation}
\frac{R}{\mu_{\text{pool}}}
\;\le\;
T_{\text{train}}
-
\frac{T_{\text{bcast}}}{\kappa}.
\label{eq:r_mu_bound}
\end{equation}

Substituting \Cref{eq:r_mu_bound} into the numerator of \Cref{eq:delta_cons} gives
\begin{equation}
T_{\text{bcast}} + \frac{R}{\mu_{\text{pool}}}
\;\le\;
T_{\text{train}}
+
\left(1 - \frac{1}{\kappa}\right) T_{\text{bcast}}.
\end{equation}

Dividing both sides by $T_{\text{train}}$ and taking the ceiling,
\begin{equation}
\left\lceil
\frac{T_{\text{bcast}} + \frac{R}{\mu_{\text{pool}}}}{T_{\text{train}}}
\right\rceil
\;\le\;
1
+
\left\lceil
\left(1 - \frac{1}{\kappa}\right)
\frac{T_{\text{bcast}}}{T_{\text{train}}}
\right\rceil.
\end{equation}

Substituting back into \Cref{eq:delta_cons} yields a tightened bound:
\begin{equation}
\Delta_{\max}^{\mathrm{cons}}
\;\le\;
\kappa
+
\left\lceil
\left(1 - \frac{1}{\kappa}\right)
\frac{T_{\text{bcast}}}{T_{\text{train}}}
\right\rceil.
\label{eq:delta_overlap}
\end{equation}

\subsection{Implication for $\kappa=2$}

For the common case $\kappa=2$, the overlap condition implies
$T_{\text{bcast}} < 2T_{\text{train}}$, and thus
\[
0
<
\frac{1}{2}\frac{T_{\text{bcast}}}{T_{\text{train}}}
<
1.
\]
Therefore,
\begin{equation}
\Delta_{\max}^{\mathrm{cons}} \;\le\; 3.
\end{equation}

This bound corresponds to a worst-case execution in which new-policy rollouts
become available only after dissemination completes.
In practice, rollout workers begin generating trajectories as soon as they
receive the update during dissemination, making the observed staleness
typically smaller than this bound.

\paragraph{Corollary: Single-Parameter Configuration.}
Consider the configuration used by \sysname{}, in which the publication period
is set to $\kappa = S-1$.
Substituting into the conservative bound yields
\[
\Delta_{\max}^{\mathrm{cons}}
=
S-1
+
\left\lceil
\frac{T_{\text{bcast}}+\frac{R}{\mu_{\text{pool}}}}{T_{\text{train}}}
\right\rceil
-1.
\]

If the system satisfies the overlap condition and $T_{\text{bcast}} / T_{\text{train}} < 1$,
which holds in all our experimental settings, then
\[
\left\lceil
\frac{T_{\text{bcast}}+\frac{R}{\mu_{\text{pool}}}}{T_{\text{train}}}
\right\rceil
\le 2,
\]
and therefore
\[
\Delta_{\max}^{\mathrm{cons}} \;\le\; S.
\]
This result justifies exposing $S$ as the sole staleness control parameter in \sysname{}.

\section{\sysname{} execution}
\subsection{Overall Procedure}
We illustrate the end-to-end execution model of \sysname{} in \Cref{alg:e2e}, which only includes Rollout Plane and Learning Plane since training process is transparent to Data Plane.
\begin{algorithm}[t]
\caption{Execution of \sysname{}}
\label{alg:e2e}
\begin{algorithmic}[1]
\STATE \textbf{Shared:} replay buffer $\mathcal{B}$, worker pool $\mathcal{W}$, active set $\mathcal{A}$
\STATE \textbf{Learner state:} update index $v \leftarrow 0$
\STATE \textbf{Worker state:} each worker $i$ maintains local snapshot version $\hat{v}_i$

\STATE \textbf{Learning Plane:}
\WHILE{training not converged}
    \IF{scheduling tick or sustained capacity deviation}
        \STATE Estimate $\mu_{\text{pool}}=\sum_{i\in\mathcal{A}} a_i \mu_i$
        \STATE Compute $\mu_{\text{target}}=\gamma \cdot \mu_{\min}(\kappa)$ using \Cref{eq:capacity_requirement}
        \STATE Adjust $\mathcal{A}$ by activating/releasing workers based on $\rho_i$ (\Cref{sec:arch:scheduling})
    \ENDIF
    \IF{$\mathcal{B}$ has at least $R$ admissible trajectories with $v \ge v_t - S$}
        \STATE Sample a batch from $\mathcal{B}$ subject to staleness version
        \STATE Perform one policy update (time $\approx T_{\text{train}}$)
        \STATE $v \leftarrow v+1$
        \IF{$v \bmod \kappa = 0$}
            \STATE Publish snapshot with version $v$ and trigger dissemination (\Cref{sec:arch:p2p})
        \ENDIF
    \ENDIF
\ENDWHILE

\STATE \textbf{Rollout Plane, on each worker $i\in\mathcal{A}$ in parallel:}
\WHILE{worker $i$ is active}
    \STATE Receive and forward snapshot chunks as a relay (\Cref{sec:arch:p2p})
    \IF{a newer snapshot is fully installed}
        \STATE Update local version $\hat{v}_i \leftarrow v_{\text{new}}$
    \ENDIF
    \STATE Sample and generate $y \sim \pi_{\hat{v}_i}(\cdot\mid x)$ and compute reward $r=\mathcal{R}(x,y)$
    \STATE Push trajectory $(x,y,r,\hat{v}_i,\Omega)$ into $\mathcal{B}$
\ENDWHILE
\end{algorithmic}
\end{algorithm}

\subsection{Supplementary System Design}
\subsubsection{Low-Frequency Adjustment}
\label{app:low_frequency_schedule}
The scheduler maintains an active worker set $\mathcal{A}$ and monitors its aggregate throughput $\sum_{i\in\mathcal{A}} a_i \mu_i$.
If capacity persistently falls below $\mu_{\text{target}}$, additional workers with low unit throughput cost $\rho$ are activated; if capacity exceeds the target by a sufficient margin, expensive workers are gradually released. 
This design ensures that the learner remains saturated whenever feasible, while avoiding frequent reconfiguration and unnecessary rollout cost.

\subsection{Forwarding-Chain Construction in Peer-Assisted Broadcast}
\label{app:chain-construction}

This appendix details how \sysname{} constructs and maintains the
peer-forwarding chains used in \Cref{sec:arch:p2p}. Given a learner uplink
budget $B_0$ and a per-worker forwarding bandwidth $\hat{B}_w$, \sysname{}
instantiates
\[
N_{\text{ch}} \;\approx\; \left\lfloor \frac{B_0}{\hat{B}_w} \right\rfloor
\]
parallel chains. Each chain disseminates the full snapshot via chunked
store-and-forward streaming; every active worker is assigned to exactly one
chain in a given broadcast round and maintains one upstream parent and at
most one downstream child.

\paragraph{Worker scoring.}
Among workers selected by cost-aware activation (\Cref{sec:arch:scheduling}),
\sysname{} ranks candidates by a dynamic peer score that combines transfer
speed, success rate, and recent stability:
\[
\mathrm{Score}_i
=
\mathrm{Decay}_i \,\bigl(
0.5\,\mathrm{Speed}_i
+ 0.35\,\mathrm{Success}_i
+ 0.15\,\mathrm{Stability}_i
\bigr).
\]
The component scores are updated online from heartbeat statistics:
\begin{align*}
\mathrm{Speed}_i &= 100 \cdot \frac{\log_{10}(\mathrm{avg\_speed}_i) - 3.0}{5.0},\\
\mathrm{Success}_i &= 100 \cdot \mathrm{success\_rate}_i,\\
\mathrm{Stability}_i &=
\begin{cases}
0, & \text{if consecutive failures} > 3,\\
100\cdot \min\!\left(\mathrm{succ\_count}_i / 10,\, 1\right), & \text{otherwise},
\end{cases}\\
\mathrm{Decay}_i &= 1 - \min\!\left(\mathrm{elapsed}_i / 600,\, 0.3\right).
\end{align*}
The $\mathrm{Speed}_i$ term saturates at $100$ when the measured forwarding
rate reaches the per-worker cap (e.g., $100$\,Mbps in our experiments).
$\mathrm{Decay}_i$ down-weights stale measurements, so workers re-enter the
ranking smoothly after periods of inactivity.

\paragraph{Chain assembly.}
The top-$N_{\text{ch}}$ workers under this ranking are designated as
first-hop seeds, allowing them to receive snapshot chunks directly from the
learner and start generating rollouts as early as possible. The remaining
active workers are appended to the chain tails in rank order, in a simple
round-robin manner across the $N_{\text{ch}}$ chains. The forwarding
topology is therefore dynamic rather than static: workers admitted in later
rounds are integrated into chains based on their current scores, and workers
exhibiting degraded transfer quality are progressively deprioritized
(seeds first, then moved toward chain tails, then dropped from the
forwarding role entirely while remaining as rollout-only workers).

\paragraph{Coefficient choices.}
The weights $(0.5, 0.35, 0.15)$, the failure threshold of $3$, the success
saturation count of $10$, and the decay window of $600$ seconds are
operational defaults rather than tuned hyperparameters; in our experiments
ranking quality is dominated by $\mathrm{Speed}_i$ and the system is robust
to moderate perturbations of these constants.

\section{Fault Tolerance under Single-Worker Failure}
\label{app:fault-tolerance}

\sysname{}'s default deployment provisions a small redundancy buffer
($\gamma = 1.1$ in our experiments, see \Cref{sec:arch:scheduling}), which
absorbs typical short-term throughput fluctuations without triggering any
reconfiguration. To stress-test the system under more adversarial
conditions, we additionally evaluate a no-redundancy setting ($\gamma = 1.0$)
in which $8$ rollout workers exactly saturate the learner. In this regime,
the loss of a single worker corresponds to a $1/8 = 12.5\%$ capacity loss,
which is large enough to expose any pipeline stall behavior.

\paragraph{Recovery mechanism.}
\sysname{} continuously monitors per-worker liveness and progress through
the same heartbeat statistics used for chain ranking
(Appendix \ref{app:chain-construction}). When effective capacity drops below
$\mu_{\text{target}}$, the scheduler launches a replacement worker drawn
from the standby pool on the same compute platform; the replacement
retrieves the current snapshot from cloud storage and rejoins the active
set. For broadcast-stage failures specifically, the affected forwarding
chain is rebuilt: the failed worker is excised, its downstream tail is
reattached to the nearest healthy parent, and the replacement is appended
to the chain in rank order.

\paragraph{Fault injection results.}
\Cref{tab:fault-injection} reports the impact of single-worker faults
injected separately during broadcast and rollout stages. Across all four
scenarios \sysname{} recovers to full saturation within a bounded window
(240--312\,s, equivalent to roughly $0.15$--$0.20$ of one training step at
$T_{\text{train}} \approx 1500$\,s for our 8B setting), with bubble-ratio
increases below $3\%$.

\begin{table}[h]
\centering
\small
\caption{Single-worker fault injection in a minimally saturated 8-worker
deployment ($\gamma = 1.0$). Throughput drop is the peak instantaneous
reduction; recovery time is measured from fault injection to the moment
$\mu_{\text{pool}}$ returns to $\mu_{\text{target}}$; bubble-ratio increase
is computed against the fault-free baseline.}
\label{tab:fault-injection}
\begin{tabular}{lccc}
\toprule
\textbf{Fault scenario} & \textbf{Throughput drop} & \textbf{Recovery time} & \textbf{Bubble-ratio increase} \\
\midrule
Broadcast: 1 forwarder slows $0.5\times$ & $8.94\%$ & $312$\,s & $1.56\%$ \\
Broadcast: 1 forwarder drops out & $16.45\%$ & $312$\,s & $2.63\%$ \\
Rollout: 1 worker slows $0.5\times$ & $6.80\%$ & $240$\,s & $0.97\%$ \\
Rollout: 1 worker drops out & $12.50\%$ & $240$\,s & $1.88\%$ \\
\bottomrule
\end{tabular}
\end{table}

\paragraph{Why broadcast-stage faults appear more severe than rollout-stage
faults.}
At first glance the broadcast-stage drop-out result ($16.45\%$) is larger
than the rollout-stage drop-out result ($12.50\% = 1/8$). The gap reflects
the cascading effect of a failed forwarder: when a relay $\mathcal{A}_{k,j}$
on chain $j$ disappears, every downstream worker on the same chain
temporarily cannot install the new snapshot, so the entire downstream
segment falls back to generating rollouts under its previous policy version
until the chain is rebuilt and the new snapshot propagates through. During
this transient, the per-step trajectory throughput observed by the learner
is reduced by more than the $1/8$ floor implied by losing a single worker.
In contrast, a rollout-stage fault removes exactly one worker's contribution
and has no effect on other workers' policy installation, so the peak drop
matches the per-worker share. The same cascading effect explains why
broadcast recovery ($312$\,s) takes longer than rollout recovery ($240$\,s):
chain reconstruction must complete and the snapshot must traverse the
rebuilt segment before downstream workers can resume generation under the
new version, whereas a rollout-stage replacement only needs to download the
snapshot once and start generating.

\paragraph{Operating-regime caveat.}
These results characterize the worst case under the no-redundancy setting
($\gamma = 1.0$). The default $\gamma = 1.1$ used in our main experiments
absorbs single-worker faults entirely without requiring replacement, at the
cost of a small ($\sim 10\%$) overprovisioning. We do not claim robustness
to large-scale concurrent failures (e.g., $\ge 50\%$ unavailable workers),
which fall outside the operating regime targeted by \sysname{}.

\section{Supplementary Experiments}
\subsection{Results of Qwen3-4B}
\label{app:result_4b}
\begin{figure}[t]
  \centering
  \begin{minipage}{0.48\linewidth}
    \centering
    \includegraphics[width=\linewidth]{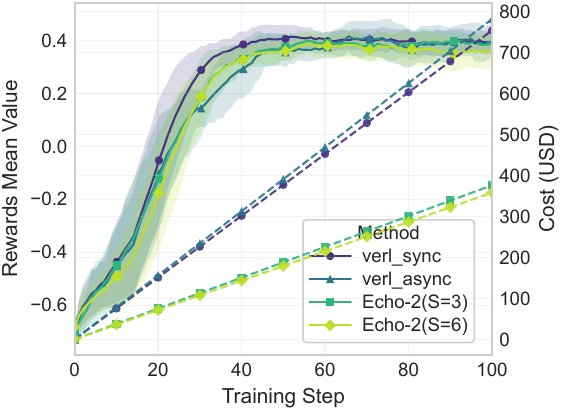}
    \caption{Cost--quality on AIME for Qwen3-4B.}
    \label{fig:costquality_qwen3_4b}
  \end{minipage}
  \hfill
  \begin{minipage}{0.48\linewidth}
    \centering
    \includegraphics[width=\linewidth]{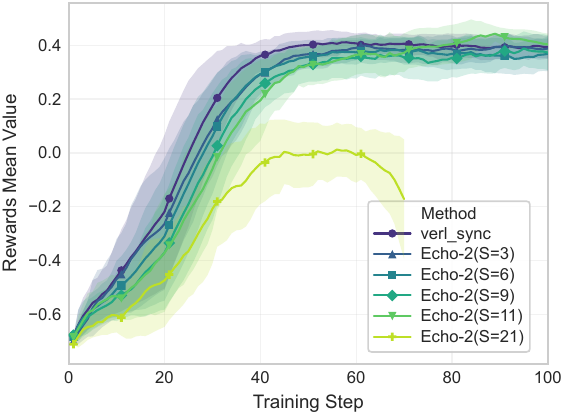}
    \caption{Effect of bounded staleness $S$ on RL quality in \sysname{} for Qwen3-4B.}
    \label{fig:staleness_qwen3_4b}
  \end{minipage}
\end{figure}

In this section, we show cost-quality comparison for Qwen3-4B, which demonstrates similar performance to Qwen3-8B in \Cref{fig:mainresults}. 
Moreover, as shown in \Cref{fig:staleness_qwen3_4b}, we also conduct empirical experiments of staleness for Qwen3-4B with standard GRPO.

\subsection{Wide-Range Benchmarks}
\label{app:extra_bench}

\Cref{tab:rl_reward_math} reports reward scores after RL post-training on 5 math reasoning benchmarks:
AIME24~\cite{maa2024aime}, OmniMath~\cite{gao2024omni}, JEE~\cite{arora2023have}, HardMath~\cite{fan2024hardmath}, and \texttt{IMO-answer-400}~\cite{luong2025towards}.
We compare \sysname{} ($S=3$) with \verl{} under the same reward model and training configuration, using Qwen3-4B and Qwen3-8B as the base models. Across all datasets and both model scales, \sysname{} maintains reward performance comparable to \verl{}, demonstrating that distributed rollouts with bounded staleness do not degrade RL optimization quality and provide a cost-efficiency opportunity, and \sysname{} realizes it.

\begin{table}[t]
\centering
\small
\caption{Reward scores after RL post-training on math reasoning benchmarks. AIME24 reports avg@64, JEE reports avg@8, and OmniMath / HardMath /IMO-A report avg@1 (i.e., Pass@1). IMO-A denotes \texttt{IMO-answer-400}. All results are reported under the same training configuration.}
\setlength{\tabcolsep}{5pt}
\begin{tabular}{lccccccc}
\toprule
\textbf{Method} & \textbf{Model} 
& \textbf{AIME24} 
& \textbf{OmniMath} 
& \textbf{JEE} 
& \textbf{HardMath} 
& \textbf{IMO-A} 
& \textbf{MEAN} \\
\midrule
\multirow{2}{*}{\ initial{}} 
 & Qwen3-4B 
 & 25.0 & 28.5 & 20.15 & 11.15 & 11.0 & 19.16 \\
 & Qwen3-8B 
 & 29.1 & 28.12 & 18.57 & 11.18 & 11.0 & 19.59 \\
\midrule

\multirow{2}{*}{\verl{}} 
 & Qwen3-4B 
 & 45.78 & 40.65 & 36.82 & 24.17 & 23.25 & 34.13 \\
 & Qwen3-8B 
 & 47.92 & 41.92 & 32.31 & 25.33 & 29.0 & 35.30 \\
\midrule
\multirow{2}{*}{\sysname{}} 
 & Qwen3-4B 
 & 45.16 & 41.67 & 34.32 & 26.3 & 20.75 & 33.64 \\
 & Qwen3-8B 
 & 48.8 & 40.31 & 39.51 & 26.87 & 23.25 & 35.75 \\
\bottomrule
\end{tabular}
\label{tab:rl_reward_math}
\end{table}

\subsection{Validation on a Larger Model and Comparison with Prime-RL}
\label{app:32b-prime-rl}

To assess whether \sysname{}'s design generalizes beyond the 4B/8B regime
reported in the main paper, and to compare directly against a recent
decentralized-rollout system, we run \sysname{} on a 32B-parameter base
model and benchmark it against Prime-RL~\citep{team2025intellect} under
matched hyperparameters.

\paragraph{Setup.}
We post-train QwQ-32B with GRPO ($\epsilon = 0.2$, $\delta = 4$) on the
\textsc{Target-Short} task from~\citet{team2025intellect} for $100$ training
steps, with $256 \times 16$ samples per rollout step. The average response
length is approximately $4{,}000$ tokens for both systems. All RL
hyperparameters (clipping range, KL coefficient, batch size, sampling
temperature) are matched across systems.

For training, both systems use $8\times$H100 GPUs. The two systems differ
only in their rollout deployment: Prime-RL runs $10\times$H100 rollout
workers on a single intranet node with effectively unlimited bandwidth
between training and rollout, representing its idealized upper bound.
\sysname{} runs $76\times$RTX~5090 rollout workers connected over a
realistic wide-area network, served by Parallax~\citep{tong2025parallax}.

\paragraph{Cost comparison.}
\Cref{tab:prime-rl-comparison} reports total dollar cost over the $100$-step
run, computed using the same public rental prices listed in
\Cref{tab:cost_sources} ($\$3.06$/hr for A100, $\$0.35$/hr for RTX~5090; we
use $\$2.90$/hr as a representative rate for H100 rollout workers, which
is the lower end of common cloud quotes at the time of writing). Despite
operating under the more constrained wide-area regime, \sysname{} achieves
a $4.6\%$ end-to-end cost reduction relative to Prime-RL's idealized
intranet deployment. The cost gap arises because \sysname{}'s overlap-aware
provisioning and cost-aware activation make commodity RTX~5090 workers a
viable substitute for high-end H100 rollout GPUs without sacrificing
learner saturation; the same substitution is not available to Prime-RL,
whose architecture does not expose unit-throughput cost $\rho_i = c_i/\mu_i$
as a scheduling input.

\begin{table}[h]
\centering
\small
\caption{End-to-end cost comparison on QwQ-32B over $100$ GRPO training
steps with matched hyperparameters. Prime-RL is evaluated on an idealized
intranet deployment; \sysname{} is evaluated under realistic wide-area
bandwidth.}
\label{tab:prime-rl-comparison}
\begin{tabular}{lccc}
\toprule
& \textbf{Training GPUs} & \textbf{Rollout GPUs} & \textbf{Total cost (\$)} \\
\midrule
Prime-RL          & $8 \times$ H100 & $10 \times$ H100      & $1{,}914$ \\
\sysname{} (ours) & $8 \times$ H100 & $76 \times$ RTX~5090 & $1{,}826$ \\
\bottomrule
\end{tabular}
\end{table}

\paragraph{Interpreting the magnitude of the gap.}
The $4.6\%$ improvement over Prime-RL is substantially smaller than the
$33$--$36\%$ improvement \sysname{} achieves over the centralized
\textsc{verl} baseline (\Cref{sec:exp:costquality}), and the two numbers
should not be conflated. The comparison against \textsc{verl} measures the
total cost saving from moving rollout generation off centrally managed
data center GPUs onto distributed commodity workers and combines two
independent gains: (i) the architectural shift to disaggregated rollout
execution and (ii) the system mechanisms (overlap-aware provisioning,
peer-forwarding broadcast, cost-aware activation) that make this shift
feasible under wide-area constraints. The comparison against Prime-RL,
which already adopts disaggregated rollout, isolates the contribution of
component (ii) alone. The $4.6\%$ figure is also conservative because
Prime-RL is evaluated on an idealized intranet, which is the most favorable
setting for any centralized-broadcast design; in a wide-area deployment
with a finite uplink budget at the trainer, Prime-RL's broadcast cost would
grow with rollout fleet size while \sysname{}'s peer-forwarding cost
remains bounded (\Cref{sec:exp:broadcast}).

\paragraph{Training stability under chunked dissemination.}
A natural concern at the 32B scale is whether the longer per-snapshot
broadcast time interacts poorly with chunked policy dissemination and
destabilizes optimization. Over the full $100$-step run we observe no
such effect: reward trajectories track the Prime-RL baseline closely
throughout training, and final reward at step $100$ matches the Prime-RL
endpoint within run-to-run variance. This is consistent with the system
design, which decouples \emph{how} a snapshot is delivered from
\emph{which} snapshot version a rollout is tagged with: chunked
dissemination only affects when a worker installs the new policy, while
the staleness budget $S$ continues to bound the version gap consumed by
the learner regardless of broadcast topology.

\paragraph{Caveat on resource scope.}
We were not able to evaluate at the 70B+ scale due to compute budget. The
QwQ-32B result above is the largest model on which we have run \sysname{}
end-to-end. We expect the design to remain feasible at larger scales
because $T_{\text{train}}$ also grows with model size, preserving the
overlap region required by \Cref{eq:capacity_requirement}; we leave a
direct empirical study to future work.

\section{Data Plane Implementation and Poker Game Alignment via Sandbox Integration}
\label{sec:poker_case_study}

To demonstrate the versatility of ECHO-2's decoupled \textit{Data Plane}, we extend our evaluation from static mathematical reasoning to a dynamic, interactive environment: \textbf{No-Limit Texas Hold'em}. This case study illustrates how ECHO-2 adapts to non-standard modalities (game logs and episodic returns) \emph{without modifying} the underlying \textit{Learning Plane} or \textit{Rollout Plane}. Concretely, we only instantiate a task-specific \textbf{Data Plane Adapter} that (i) interfaces with a poker sandbox, (ii) standardizes raw logs into the canonical rollout schema, and (iii) materializes the additional metadata $\Omega$ (e.g., token masks and normalized advantages) required by GRPO, yielding the canonical record $\tau=(x,y,r,v,\Omega)$ consumed by the shared replay buffer and the learner.

\begin{figure}[h]
    \centering
    \includegraphics[width=0.6\linewidth]{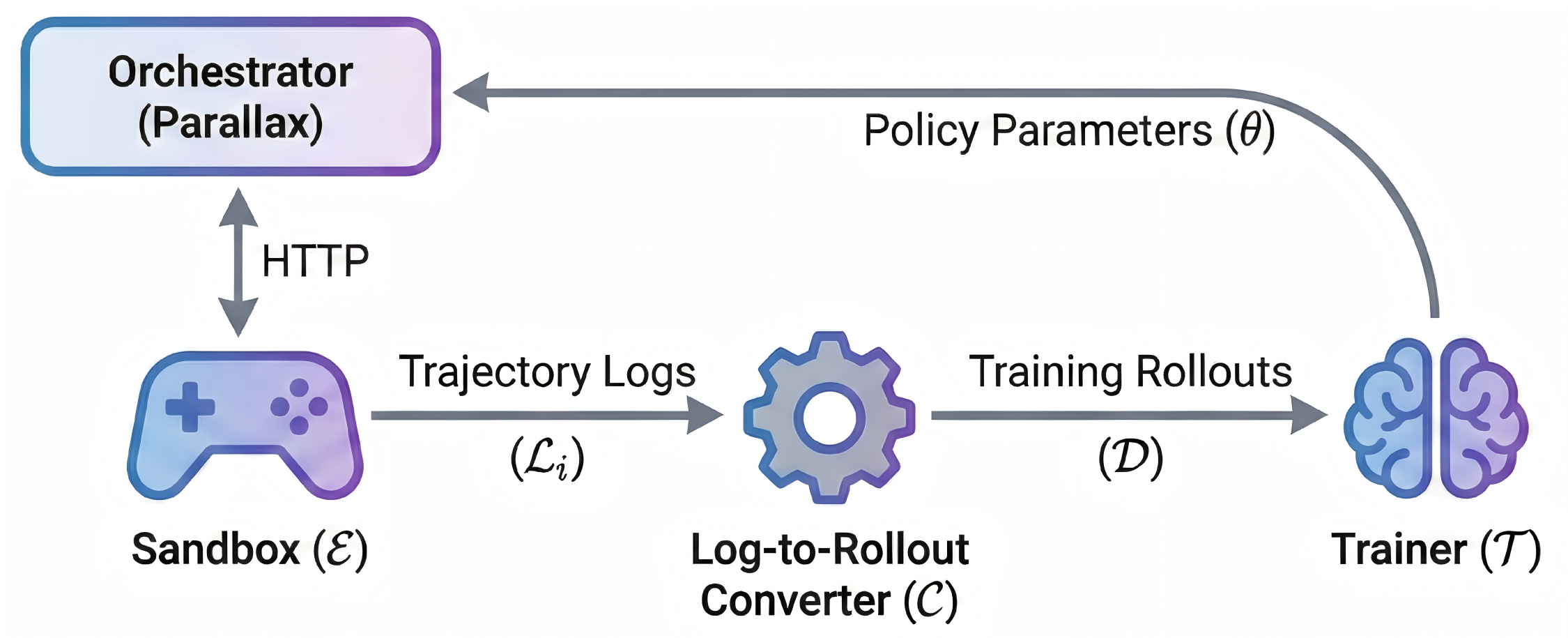}
    \caption{Overview of the Echo-2 Poker Game Alignment system. The Orchestrator (Parallax) interfaces with the Sandbox ($\mathcal{E}$) to generate Trajectory Logs ($\mathcal{L}_i$). The Log-to-Rollout Converter ($\mathcal{C}$) processes these logs into Training Rollouts ($\mathcal{D}$), which are then used by the Trainer ($\mathcal{T}$) to update the policy parameters ($\theta$), closing the iterative training loop.}
    \label{fig:sandbox}
\end{figure}

\subsection{Data Plane Interfaces and Task Integration}
\label{sec:arch:data}

The Data Plane defines the \emph{task semantics} of \sysname{} while preserving the versioned execution in \Cref{sec:arch:versioned} and the end-to-end loop in \Cref{alg:e2e}. Concretely, it specifies how a workload is mapped to immutable, version-tagged trajectory records stored in the replay buffer: $\tau \;=\; (x,\; y,\; r,\; v,\; \Omega)$, where $(x,y)$ is the prompt-response pair, $r$ is the scalar reward, $v$ is the snapshot version used to generate $y$, and $\Omega$ is optional task metadata. The buffer $\mathcal{B}$ indexes $\tau$ by version and enforces bounded staleness.

A task is integrated by implementing a Data Plane adapter that (i) constructs prompts $x$, (ii) defines the reward function $\mathcal{R}$ used by rollout workers to produce $r=\mathcal{R}(x,y)$, and (iii) defines how $\Omega$ is materialized into learner-side training signals (e.g., masks and normalized advantages) under the chosen objective (e.g., GRPO with KL regularization).

\subsection{System Overview: A Data Plane Instantiation}
We implement a specialized Data Plane Adapter that bridges the raw poker sandbox and ECHO-2's training interface. As shown in Figure~\ref{fig:sandbox}, the pipeline consists of three phases: \emph{Sandbox Interaction}, \emph{Log Standardization}, and \emph{Reward-Augmented Rollout Generation}. The adapter outputs a unified rollout tuple that can be consumed directly by the generic \textit{Rollout Plane} and \textit{Learning Plane}.

Let the policy be an autoregressive language model $\pi_\theta(\mathbf{y}\mid \mathbf{x})$, where $\mathbf{x}$ represents the serialized game context and $\mathbf{y}$ represents the agent's decision (betting action text). We denote a reference policy (for KL regularization) as $\pi_{\text{ref}}$.

\paragraph{One-line task switching via sandbox adapters.}
A key benefit of the decoupled Data Plane is that switching to a new interactive task
only requires swapping the sandbox adapter configuration, while the Rollout/Learning Planes
(and the replay schema $\tau=(x,y,r,v,\Omega)$) remain unchanged.

\begin{codebox}[title={Unified Orchestration API (Poker $\leftrightarrow$ MOBA)}]
# Task-specific: only the Data Plane adapter changes.
# Switching tasks is configuration-only:
# adapter.env = "moba://hok_v2" ; adapter.serializer = "moba_obs->prompt_v2"
adapter = SandboxAdapter(
    env="poker://nlhe_6max_v1",           # or "moba://dota2_v3", "moba://hok_v2"
    serializer="envstate->chat_prompt_v2", 
    action_renderer="action->text_v1",  
    reward_spec="episodic_return", 
    metadata_spec="grpo:turn_mask+adv_norm" 
)

echo = Echo2(policy=pi_theta, reference=pi_ref)

echo.set(dataset=adapter.stream(mode="offline_or_live"))   
echo.set(staleness=3)                                    
echo.warmup()                                          
echo.run()                                    
\end{codebox}

\subsection{Phase 1: Environment Interaction (Sandbox $\rightarrow$ Raw Logs)}
We deploy a sandbox environment $\mathcal{E}$ simulating a poker table. For each episode $i$, the environment records a raw interaction log $\mathcal{L}_i$:
\begin{equation}
\mathcal{L}_i = \{(s_{i,t}, a_{i,t}, r_{i,t})\}_{t=1}^{T_i},
\end{equation}
where:
\begin{itemize}
    \item $s_{i,t}$ is a textual description of the private hand, community cards, pot size, and derived odds (e.g., \texttt{"Hand: [Ah, Kd], Board: [Qs, Th, 2c], Pot: 100"}).
    \item $a_{i,t}$ is a structured action rendered as text (e.g., \texttt{"Action: Raise 50"}).
    \item $r_{i,t}$ is the \emph{immediate chip delta}, i.e., the change in chip stack relative to the previous turn.
\end{itemize}
Unlike math tasks where rollouts are generated by the model under training, poker logs may initially come from rule-based baselines or prior model iterations, demonstrating ECHO-2's ability to consume \emph{off-policy} data (and strictly on-policy data if connected to live rollout workers). In \sysname{}, the scalar reward $r$ is produced in the Rollout Plane (co-located with environment interaction), while the Data Plane defines $\mathcal{R}$ and the post-processing rules used to derive $\Omega$ for learning.

\subsection{Phase 2: Standardization and Conversion (Raw Logs $\rightarrow$ Canonical Messages)}
The core responsibility of the Data Plane is to convert heterogeneous logs $\mathcal{L}_i$ into a unified rollout format compatible with the generic Learning Plane. The adapter $\mathcal{C}$ transforms the raw log into a chat-formatted message sequence $\mathcal{M}_i$ by flattening complex game states into a standard prompt-response template:
\begin{equation}
\mathcal{M}_i =
\bigl[
m_{i,0}^{\text{sys}},\;
m_{i,0}^{\text{usr}},\;
m_{i,1}^{\text{asst}},\;
m_{i,1}^{\text{usr}},\;\dots,\;
m_{i,T_i}^{\text{asst}}
\bigr],
\end{equation}
where $m_{i,0}^{\text{sys}}=\texttt{SystemPrompt}$ encodes global poker rules $\mathcal{P}_{\text{rules}}$, and each turn is represented as a user state message followed by an assistant action message:
\begin{equation}
m_{i,t}^{\text{usr}}=\texttt{``State: }\!s_{i,t}\texttt{''},
\qquad
m_{i,t}^{\text{asst}}=\texttt{``}\!a_{i,t}\texttt{''}.
\end{equation}
Optionally, for bookkeeping we may insert rewards as user messages $\texttt{``Reward: }r_{i,t}\texttt{''}$; however, training signals are ultimately computed from numeric rewards inside the Data Plane.

We then linearize $\mathcal{M}_i$ using the tokenizer chat template and obtain token IDs:
\begin{equation}
\mathbf{x}_i=(x_{i,1},\dots,x_{i,L_i})=\text{Tokenize}(\mathcal{M}_i),
\end{equation}
along with an attention mask $\mathbf{a}_i\in\{0,1\}^{L_i}$. ECHO-2 uses left padding to batch variable-length episodes.

The following implementation demonstrates how raw environment outputs are iteratively converted into the user-assistant message structure:

\begin{codebox}[title={Constructing Canonical Messages}]
# Extract from GemEnvRollout.get_lm_inputs
messages = [
    {"role": "system", "content": self.system_prompt},
    {"role": "user", "content": self.prefix_lookup[env_id]}
]

for idx, content in enumerate(env_output["history"]):
    messages[-1]["content"] += f"\nTurn {idx + 1}:\n"
    
    # Flatten State s_{i,t}
    if "state" in content:
        FORMAT = "<answer> ... </answer>"
        messages[-1]["content"] += f"State:\n{content['state']}\n Always output: {FORMAT}\n"

    # Append Assistant Action a_{i,t}
    if "llm_response" in content:
        messages.append({"role": "assistant", "content": content["llm_response"]})
        
    # Optional: Insert intermediate rewards for bookkeeping
    if "reward" in content:
        messages.append({"role": "user", "content": f"Reward:\n{content['reward']}\n"})
\end{codebox}

\subsection{Phase 3: Turn-Aware Masking and Reward-Augmented Rollouts}
Poker supervision is sparse and episodic; therefore, the Data Plane additionally computes (i) \textbf{turn-aware masks} that restrict learning to assistant tokens, and (ii) \textbf{advantages} derived from final chip outcomes.

\subsubsection{Turn-Aware Masking}
We construct turn indicators using a special \emph{turn-start} token ID $\tau_{\text{start}}$ (e.g., \texttt{<|im\_start|>} in Qwen-style templates). Define:
\begin{equation}
u_{i,t}=\mathbb{I}[x_{i,t}=\tau_{\text{start}}],
\qquad
c_{i,t}=\sum_{k=1}^{t}u_{i,k},
\end{equation}
where $c_{i,t}$ is the chat-turn index of token $t$. The assistant-response mask is:
\begin{equation}
m^{\text{resp}}_{i,t}
=
\mathbb{I}[c_{i,t}>1]\cdot \mathbb{I}[c_{i,t}\bmod 2 = 1],
\end{equation}
selecting tokens after the system prompt that belong to assistant turns. We set the loss mask as $m^{\text{loss}}_{i,t}=m^{\text{resp}}_{i,t}$, so learning is restricted to the agent's action tokens. Under next-token prediction, masks are aligned with shifted targets $y_{i,t}=x_{i,t+1}$.

The implementation below corresponds to the calculation of $m^{\text{resp}}_{i,t}$ and the logic for aligning rewards to turn boundaries:

\begin{codebox}[title={Turn-Aware Mask Computation}]
def get_masks_and_scores(input_ids, tokenizer, all_scores, enable_response_mask=False):
    special_token, reward_token = get_special_tokens(tokenizer)

    # Calculate turn indicators c_{i,t}
    turn_starts = torch.where(input_ids == special_token, 1, 0)
    turn_indicators = torch.cumsum(turn_starts, dim=-1)
    
    # Generate Response Mask m^{resp}_{i,t}
    # Selects tokens where turn count is odd (assistant) and > 1 (after sys prompt)
    response_mask = (turn_indicators % 2 == 1) & (turn_indicators > 1)

    # Assign scores to the last token of the turn
    score_tensor = torch.zeros_like(input_ids, dtype=torch.float32)
    scores = [sum(i) for i in all_scores] # Sum of rewards R_i
    score_tensor[:, -1] = torch.tensor(scores, dtype=torch.float32)
    
    # Alignment adjustments for causal masking
    score_tensor = score_tensor[:, 1:] 
    loss_mask = response_mask[:, :-1] 
    
    return score_tensor, loss_mask, response_mask
\end{codebox}

\subsubsection{Outcome-Based Returns and Group-wise Normalization}
While poker is high-variance, our primary evaluation metric is the \emph{final chip change}. For episode $i$, the trajectory-level return is:
\begin{equation}
R_i=\sum_{t=1}^{T_i} r_{i,t},
\end{equation}
equal to the net profit/loss in chips. We use the trajectory-level return as the scalar reward stored in the record, i.e., $r_i := R_i$.

To reduce variance and stabilize policy updates, the Data Plane applies \textbf{group-wise normalization}. For a group of episodes $G$ (e.g., sharing similar initial private hands or other coarse state descriptors), we compute the normalized advantage:
\begin{equation}
\hat{A}_i
=
\frac{
R_i-\text{Mean}\bigl(\{R_j\}_{j\in G}\bigr)
}{
\text{Std}\bigl(\{R_j\}_{j\in G}\bigr)+\epsilon
}.
\end{equation}
We then broadcast $\hat{A}_i$ to the response tokens:
\begin{equation}
\hat{A}_{i,t} = \hat{A}_i\cdot m^{\text{resp}}_{i,t},
\end{equation}
so that only assistant tokens receive non-zero advantage.

This normalization logic supports multiple grouping strategies (e.g., by initial state or batch) to compute the standardized returns used in the GRPO objective:

\begin{codebox}[title={Group-wise Reward Normalization}]
def _normalize_score_tensor(self, score_tensor, env_outputs):
    # Grouping logic (G)
    if self.reward_normalization['grouping'] == "state":
        group_tags = [out["group_id"] for out in env_outputs]
    
    # Calculate Mean and Std for the group
    # \hat{A}_i = (R_i - Mean) / (Std + epsilon)
    norm_func = lambda x: (x - x.mean(dim=-1, keepdim=True)) / \
                          (x.std(dim=-1, keepdim=True) + 1e-6)

    # Map indices to groups
    group2index = {}
    for i, tag in enumerate(group_tags):
        if tag not in group2index: group2index[tag] = []
        group2index[tag].append(i)

    # Apply normalization per group
    acc_scores = score_tensor[:, -1]
    normalized = acc_scores.clone()
    for group, idxs in group2index.items():
        normalized[idxs] = norm_func(normalized[idxs])

    score_tensor[:, -1] = normalized
    return score_tensor
\end{codebox}

\subsubsection{GRPO-Style Policy Gradient Objective}

The adapter emits the canonical version-tagged record $\tau_i = (x_i, y_i, r_i, v_i, \Omega_i)$, where $r_i = \sum_{t=1}^{T_i} r_{i,t}$ is the episode return and $\Omega_i$ includes task metadata such as $(m^{\text{loss}}_i, m^{\text{resp}}_i)$ and grouping tags used to compute $\hat{A}_i$. After sampling $\tau_i$ from the replay buffer, the learner materializes the training tensors $(\mathbf{x}_i, \mathbf{a}_i, m^{\text{loss}}_i, m^{\text{resp}}_i, \hat{\mathbf{A}}_i)$.

To account for the distribution shift between the sampling policy $\pi_{\text{sampler}}$ and the current learner $\pi_{\text{learner}}$, we define the token-level likelihood ratio for token $t$ in episode $i$ as:
\begin{equation}
\rho_{i,t}(\theta) = \frac{\pi_\theta(y_{i,t} \mid \mathbf{x}_i, y_{i,<t})}{\pi_{\theta_{\text{old}}}(y_{i,t} \mid \mathbf{x}_i, y_{i,<t})}.
\end{equation}

The training objective $\mathcal{J}(\theta)$ incorporates truncated importance sampling to stabilize updates when reusing off-policy data from the buffer:
\begin{equation}
\mathcal{J}(\theta) = \mathbb{E}_{a \sim \pi_{\text{sampler}}(\theta_{\text{old}})} \left[ \underbrace{\min \left( \frac{\pi_{\text{learner}}(a, \theta_{\text{old}})}{\pi_{\text{sampler}}(a, \theta_{\text{old}})}, C \right)}_{\text{truncated importance ratio}} \cdot \bar{\mathcal{J}}(\theta) \right],
\end{equation}
where $C$ is a hyper-parameter and $\bar{\mathcal{J}}(\theta)$ denotes the GRPO-style clipped surrogate objective with KL regularization:
\begin{equation}
\bar{\mathcal{J}}(\theta) = \frac{1}{\sum_t m^{\text{resp}}_{i,t}} \sum_{t=1}^{L_i-1} m^{\text{resp}}_{i,t} \cdot \min \Bigl( \rho_{i,t}(\theta)\hat{A}_{i,t}, \, \text{clip}\bigl(\rho_{i,t}(\theta), 1-\epsilon_c, 1+\epsilon_c\bigr)\hat{A}_{i,t} \Bigr) - \beta \, D_{\text{KL}}(\pi_\theta \| \pi_{\text{ref}}).
\end{equation}

Intuitively, the truncated ratio prevents gradient instability when the current policy deviates significantly from the data-collection policy. This allows the Learning Plane to robustly leverage diverse experiences from the Data Plane, demonstrating that poker environment support requires only a specialized Data Plane instantiation.

\subsection{Texas Hold'em Performance}
\label{sec:poker_analysis}

\begin{table}[ht]
\centering
\caption{Texas Hold'em evaluation of different player policies against three rule-based opponents and an LLM opponent (highlighted). The reported metric is the \emph{final chip change} (net chips at the end of an episode/match relative to the initial stack). Positive values indicate net profit, while negative values indicate net loss. Only Qwen3-0.6B includes a GRPO-trained variant (second row, marked +GRPO); all other rows are direct LLM policies without GRPO training.}
\small
\label{tab:poker}
\begin{tabular}{lcccc}
\toprule
Model & Rule-based$_1$ & Rule-based$_2$ & Rule-based$_3$ & \cellcolor{blue!10}LLM \\
\midrule
Qwen3-0.6B~\citep{qwen3technicalreport} & 0.571  & 0.514  & 0.592 & \cellcolor{blue!10}-1.677\\
Qwen3-0.6B~\citep{qwen3technicalreport} {\color{blue!60}\scriptsize +GRPO} & -0.195 & -0.599 & -0.451 & \cellcolor{blue!10}1.245 \\
\midrule
Qwen3-30B-A3B~\citep{qwen3technicalreport} & -0.397 & 0.093 & -0.1225 & \cellcolor{blue!10}0.4265 \\
Qwen3-next-80B-A3B-instruct~\citep{qwen3technicalreport} & -0.399 & 0.1005 & -0.0055 & \cellcolor{blue!10}0.304 \\
GPT-5~\citep{openai2025gpt5} & -0.216 & -0.1955 & 0.1455 & \cellcolor{blue!10}0.266 \\
Grok-4~\citep{xai2025grok4} & -0.3915 & -0.2915 & -0.228 & \cellcolor{blue!10}0.911 \\
Claude-sonnet-4.5~\citep{anthropic2025claude45} & -0.1155 & -0.055 & 0.056 & \cellcolor{blue!10}0.1145 \\
Gemini-2.5-flash~\citep{comanici2025gemini25pushingfrontier} & -0.173 & -0.2165 & 0.0305 & \cellcolor{blue!10}0.359 \\
Gemini-2.5-pro~\citep{comanici2025gemini25pushingfrontier} & -0.199 & -0.2055 & -0.04 & \cellcolor{blue!10}0.4445 \\
\bottomrule
\end{tabular}
\end{table}
We evaluate our method on a Texas Hold'em environment using the \emph{final chip change} (net chips at the end of a match relative to the initial stack; higher is better). 
In Table~\ref{tab:poker}, \textbf{columns correspond to different opponents}---three rule-based opponents (Rule-based$_1$--Rule-based$_3$) and one LLM opponent (highlighted). 
\textbf{Rows correspond to the evaluated player/agent}. 
For \textbf{Qwen3-0.6B}, we report both the \emph{base} model (first row) and its \emph{GRPO-trained} variant (second row, marked +GRPO). 
For all other backbones, we report the performance of their \emph{direct LLM policies} (i.e., without GRPO training) under the same evaluation protocol.

\paragraph{Main result: GRPO improves Qwen3-0.6B against the LLM opponent.}
For Qwen3-0.6B, GRPO flips the outcome against the LLM opponent from a net loss ($-1.677$) to a net profit ($+1.245$), demonstrating that GRPO can substantially improve end-of-game profitability in the most challenging setting. 
Meanwhile, performance against the three rule-based opponents becomes negative after GRPO, suggesting a trade-off that may be addressed by multi-opponent training or more diverse opponent sampling.


\newpage
\section*{NeurIPS Paper Checklist}

\begin{enumerate}

\item {\bf Claims}
    \item[] Question: Do the main claims made in the abstract and introduction accurately reflect the paper's contributions and scope?
    \item[] Answer: \answerYes{} 
    \item[] Justification: The abstract and introduction summarize the main contributions of ECHO-2, including centralized learning with distributed rollouts, bounded staleness, overlap-based provisioning, peer-assisted broadcast, and cost-aware worker activation. The experimental claims are GRPO post-training of 4B~32B models under the evaluated wide-area bandwidth constraints.
    \item[] Guidelines:
    \begin{itemize}
        \item The answer \answerNA{} means that the abstract and introduction do not include the claims made in the paper.
        \item The abstract and/or introduction should clearly state the claims made, including the contributions made in the paper and important assumptions and limitations. A \answerNo{} or \answerNA{} answer to this question will not be perceived well by the reviewers. 
        \item The claims made should match theoretical and experimental results, and reflect how much the results can be expected to generalize to other settings. 
        \item It is fine to include aspirational goals as motivation as long as it is clear that these goals are not attained by the paper. 
    \end{itemize}

\item {\bf Limitations}
    \item[] Question: Does the paper discuss the limitations of the work performed by the authors?
    \item[] Answer: \answerYes{} 
    \item[] Justification: \Cref{sec:limitations} discusses limitations, including the lack of formal convergence guarantees under stale rollouts, task-dependent robustness to staleness, and the current focus on a single centralized learner with distributed rollout workers.
    \item[] Guidelines:
    \begin{itemize}
        \item The answer \answerNA{} means that the paper has no limitation while the answer \answerNo{} means that the paper has limitations, but those are not discussed in the paper. 
        \item The authors are encouraged to create a separate ``Limitations'' section in their paper.
        \item The paper should point out any strong assumptions and how robust the results are to violations of these assumptions (e.g., independence assumptions, noiseless settings, model well-specification, asymptotic approximations only holding locally). The authors should reflect on how these assumptions might be violated in practice and what the implications would be.
        \item The authors should reflect on the scope of the claims made, e.g., if the approach was only tested on a few datasets or with a few runs. In general, empirical results often depend on implicit assumptions, which should be articulated.
        \item The authors should reflect on the factors that influence the performance of the approach. For example, a facial recognition algorithm may perform poorly when image resolution is low or images are taken in low lighting. Or a speech-to-text system might not be used reliably to provide closed captions for online lectures because it fails to handle technical jargon.
        \item The authors should discuss the computational efficiency of the proposed algorithms and how they scale with dataset size.
        \item If applicable, the authors should discuss possible limitations of their approach to address problems of privacy and fairness.
        \item While the authors might fear that complete honesty about limitations might be used by reviewers as grounds for rejection, a worse outcome might be that reviewers discover limitations that aren't acknowledged in the paper. The authors should use their best judgment and recognize that individual actions in favor of transparency play an important role in developing norms that preserve the integrity of the community. Reviewers will be specifically instructed to not penalize honesty concerning limitations.
    \end{itemize}

\item {\bf Theory assumptions and proofs}
    \item[] Question: For each theoretical result, does the paper provide the full set of assumptions and a complete (and correct) proof?
    \item[] Answer: \answerYes{} 
    \item[] Justification: The paper states the assumptions behind the overlap condition and the conservative staleness bound in \Cref{sec:overlap_condition} and provides the derivation in \ref{app:staleness}. We emphasize that these results are system-level provisioning bounds rather than convergence guarantees for RL optimization.
    \item[] Guidelines:
    \begin{itemize}
        \item The answer \answerNA{} means that the paper does not include theoretical results. 
        \item All the theorems, formulas, and proofs in the paper should be numbered and cross-referenced.
        \item All assumptions should be clearly stated or referenced in the statement of any theorems.
        \item The proofs can either appear in the main paper or the supplemental material, but if they appear in the supplemental material, the authors are encouraged to provide a short proof sketch to provide intuition. 
        \item Inversely, any informal proof provided in the core of the paper should be complemented by formal proofs provided in appendix or supplemental material.
        \item Theorems and Lemmas that the proof relies upon should be properly referenced. 
    \end{itemize}

    \item {\bf Experimental result reproducibility}
    \item[] Question: Does the paper fully disclose all the information needed to reproduce the main experimental results of the paper to the extent that it affects the main claims and/or conclusions of the paper (regardless of whether the code and data are provided or not)?
    \item[] Answer: \answerYes{} 
    \item[] Justification: \Cref{sec:exp:setup}--\Cref{sec:exp:ablation} describe the models, GRPO hyperparameters, hardware deployment, network regimes, baselines, cost metrics, and ablations used for the main results. Additional implementation and system details are provided in the appendices to support reproducibility.
    \item[] Guidelines:
    \begin{itemize}
        \item The answer \answerNA{} means that the paper does not include experiments.
        \item If the paper includes experiments, a \answerNo{} answer to this question will not be perceived well by the reviewers: Making the paper reproducible is important, regardless of whether the code and data are provided or not.
        \item If the contribution is a dataset and\slash or model, the authors should describe the steps taken to make their results reproducible or verifiable. 
        \item Depending on the contribution, reproducibility can be accomplished in various ways. For example, if the contribution is a novel architecture, describing the architecture fully might suffice, or if the contribution is a specific model and empirical evaluation, it may be necessary to either make it possible for others to replicate the model with the same dataset, or provide access to the model. In general. releasing code and data is often one good way to accomplish this, but reproducibility can also be provided via detailed instructions for how to replicate the results, access to a hosted model (e.g., in the case of a large language model), releasing of a model checkpoint, or other means that are appropriate to the research performed.
        \item While NeurIPS does not require releasing code, the conference does require all submissions to provide some reasonable avenue for reproducibility, which may depend on the nature of the contribution. For example
        \begin{enumerate}
            \item If the contribution is primarily a new algorithm, the paper should make it clear how to reproduce that algorithm.
            \item If the contribution is primarily a new model architecture, the paper should describe the architecture clearly and fully.
            \item If the contribution is a new model (e.g., a large language model), then there should either be a way to access this model for reproducing the results or a way to reproduce the model (e.g., with an open-source dataset or instructions for how to construct the dataset).
            \item We recognize that reproducibility may be tricky in some cases, in which case authors are welcome to describe the particular way they provide for reproducibility. In the case of closed-source models, it may be that access to the model is limited in some way (e.g., to registered users), but it should be possible for other researchers to have some path to reproducing or verifying the results.
        \end{enumerate}
    \end{itemize}

\item {\bf Open access to data and code}
    \item[] Question: Does the paper provide open access to the data and code, with sufficient instructions to faithfully reproduce the main experimental results, as described in supplemental material?
    \item[] Answer: \answerNo{} 
    \item[] Justification: The current submission does not include an anonymized public release of the ECHO-2 codebase or reproduction scripts. We describe the experimental setup in the paper and plan to release code and scripts upon publication, subject to anonymization and infrastructure constraints.
    \item[] Guidelines:
    \begin{itemize}
        \item The answer \answerNA{} means that paper does not include experiments requiring code.
        \item Please see the NeurIPS code and data submission guidelines (\url{https://neurips.cc/public/guides/CodeSubmissionPolicy}) for more details.
        \item While we encourage the release of code and data, we understand that this might not be possible, so \answerNo{} is an acceptable answer. Papers cannot be rejected simply for not including code, unless this is central to the contribution (e.g., for a new open-source benchmark).
        \item The instructions should contain the exact command and environment needed to run to reproduce the results. See the NeurIPS code and data submission guidelines (\url{https://neurips.cc/public/guides/CodeSubmissionPolicy}) for more details.
        \item The authors should provide instructions on data access and preparation, including how to access the raw data, preprocessed data, intermediate data, and generated data, etc.
        \item The authors should provide scripts to reproduce all experimental results for the new proposed method and baselines. If only a subset of experiments are reproducible, they should state which ones are omitted from the script and why.
        \item At submission time, to preserve anonymity, the authors should release anonymized versions (if applicable).
        \item Providing as much information as possible in supplemental material (appended to the paper) is recommended, but including URLs to data and code is permitted.
    \end{itemize}

\item {\bf Experimental setting/details}
    \item[] Question: Does the paper specify all the training and test details (e.g., data splits, hyperparameters, how they were chosen, type of optimizer) necessary to understand the results?
    \item[] Answer: \answerYes{} 
    \item[] Justification: \Cref{sec:exp:setup} specifies the evaluated models, GRPO settings, rollout parameters, task and reward function, hardware deployment, network regimes, baselines, and cost metrics. Additional configuration details are provided in the appendix.
    \item[] Guidelines:
    \begin{itemize}
        \item The answer \answerNA{} means that the paper does not include experiments.
        \item The experimental setting should be presented in the core of the paper to a level of detail that is necessary to appreciate the results and make sense of them.
        \item The full details can be provided either with the code, in appendix, or as supplemental material.
    \end{itemize}

\item {\bf Experiment statistical significance}
    \item[] Question: Does the paper report error bars suitably and correctly defined or other appropriate information about the statistical significance of the experiments?
    \item[] Answer: \answerYes{} 
    \item[] Justification: The main figures report smoothed reward curves and per-step variability.
    \item[] Guidelines:
    \begin{itemize}
        \item The answer \answerNA{} means that the paper does not include experiments.
        \item The authors should answer \answerYes{} if the results are accompanied by error bars, confidence intervals, or statistical significance tests, at least for the experiments that support the main claims of the paper.
        \item The factors of variability that the error bars are capturing should be clearly stated (for example, train/test split, initialization, random drawing of some parameter, or overall run with given experimental conditions).
        \item The method for calculating the error bars should be explained (closed form formula, call to a library function, bootstrap, etc.)
        \item The assumptions made should be given (e.g., Normally distributed errors).
        \item It should be clear whether the error bar is the standard deviation or the standard error of the mean.
        \item It is OK to report 1-sigma error bars, but one should state it. The authors should preferably report a 2-sigma error bar than state that they have a 96\% CI, if the hypothesis of Normality of errors is not verified.
        \item For asymmetric distributions, the authors should be careful not to show in tables or figures symmetric error bars that would yield results that are out of range (e.g., negative error rates).
        \item If error bars are reported in tables or plots, the authors should explain in the text how they were calculated and reference the corresponding figures or tables in the text.
    \end{itemize}

\item {\bf Experiments compute resources}
    \item[] Question: For each experiment, does the paper provide sufficient information on the computer resources (type of compute workers, memory, time of execution) needed to reproduce the experiments?
    \item[] Answer: \answerYes{} 
    \item[] Justification: \Cref{sec:exp:setup} reports the hardware used for the learner and rollout workers, including A100 80GB GPUs and RTX 5090 rollout workers, and \Cref{sec:exp:costquality} reports per-step training times used for cost estimation. We also report GPU rental prices and cost computation in the experimental section.
    \item[] Guidelines:
    \begin{itemize}
        \item The answer \answerNA{} means that the paper does not include experiments.
        \item The paper should indicate the type of compute workers CPU or GPU, internal cluster, or cloud provider, including relevant memory and storage.
        \item The paper should provide the amount of compute required for each of the individual experimental runs as well as estimate the total compute. 
        \item The paper should disclose whether the full research project required more compute than the experiments reported in the paper (e.g., preliminary or failed experiments that didn't make it into the paper). 
    \end{itemize}
    
\item {\bf Code of ethics}
    \item[] Question: Does the research conducted in the paper conform, in every respect, with the NeurIPS Code of Ethics \url{https://neurips.cc/public/EthicsGuidelines}?
    \item[] Answer: \answerYes{} 
    \item[] Justification: To the best of our knowledge, the research conforms to the NeurIPS Code of Ethics. The experiments use public models, public benchmarks, and system-level measurements, and do not involve private user data or human-subject studies.
    \item[] Guidelines:
    \begin{itemize}
        \item The answer \answerNA{} means that the authors have not reviewed the NeurIPS Code of Ethics.
        \item If the authors answer \answerNo, they should explain the special circumstances that require a deviation from the Code of Ethics.
        \item The authors should make sure to preserve anonymity (e.g., if there is a special consideration due to laws or regulations in their jurisdiction).
    \end{itemize}

\item {\bf Broader impacts}
    \item[] Question: Does the paper discuss both potential positive societal impacts and negative societal impacts of the work performed?
    \item[] Answer: \answerNo{} 
    \item[] Justification: The current paper focuses on system design and does not include a dedicated broader-impact discussion. Potential impacts include improved accessibility of RL post-training as well as risks from cheaper training of stronger models and distributed rollout execution.
    \item[] Guidelines:
    \begin{itemize}
        \item The answer \answerNA{} means that there is no societal impact of the work performed.
        \item If the authors answer \answerNA{} or \answerNo, they should explain why their work has no societal impact or why the paper does not address societal impact.
        \item Examples of negative societal impacts include potential malicious or unintended uses (e.g., disinformation, generating fake profiles, surveillance), fairness considerations (e.g., deployment of technologies that could make decisions that unfairly impact specific groups), privacy considerations, and security considerations.
        \item The conference expects that many papers will be foundational research and not tied to particular applications, let alone deployments. However, if there is a direct path to any negative applications, the authors should point it out. For example, it is legitimate to point out that an improvement in the quality of generative models could be used to generate Deepfakes for disinformation. On the other hand, it is not needed to point out that a generic algorithm for optimizing neural networks could enable people to train models that generate Deepfakes faster.
        \item The authors should consider possible harms that could arise when the technology is being used as intended and functioning correctly, harms that could arise when the technology is being used as intended but gives incorrect results, and harms following from (intentional or unintentional) misuse of the technology.
        \item If there are negative societal impacts, the authors could also discuss possible mitigation strategies (e.g., gated release of models, providing defenses in addition to attacks, mechanisms for monitoring misuse, mechanisms to monitor how a system learns from feedback over time, improving the efficiency and accessibility of ML).
    \end{itemize}
    
\item {\bf Safeguards}
    \item[] Question: Does the paper describe safeguards that have been put in place for responsible release of data or models that have a high risk for misuse (e.g., pre-trained language models, image generators, or scraped datasets)?
    \item[] Answer: \answerNA{} 
    \item[] Justification: The paper does not release a new pretrained language model, image generator, or scraped dataset that would require misuse-specific release safeguards. The work focuses on a system framework for distributed rollout execution.
    \item[] Guidelines:
    \begin{itemize}
        \item The answer \answerNA{} means that the paper poses no such risks.
        \item Released models that have a high risk for misuse or dual-use should be released with necessary safeguards to allow for controlled use of the model, for example by requiring that users adhere to usage guidelines or restrictions to access the model or implementing safety filters. 
        \item Datasets that have been scraped from the Internet could pose safety risks. The authors should describe how they avoided releasing unsafe images.
        \item We recognize that providing effective safeguards is challenging, and many papers do not require this, but we encourage authors to take this into account and make a best faith effort.
    \end{itemize}

\item {\bf Licenses for existing assets}
    \item[] Question: Are the creators or original owners of assets (e.g., code, data, models), used in the paper, properly credited and are the license and terms of use explicitly mentioned and properly respected?
    \item[] Answer: \answerYes{} 
    \item[] Justification:  Original owners of assets are property credited in the reference.
    \item[] Guidelines:
    \begin{itemize}
        \item The answer \answerNA{} means that the paper does not use existing assets.
        \item The authors should cite the original paper that produced the code package or dataset.
        \item The authors should state which version of the asset is used and, if possible, include a URL.
        \item The name of the license (e.g., CC-BY 4.0) should be included for each asset.
        \item For scraped data from a particular source (e.g., website), the copyright and terms of service of that source should be provided.
        \item If assets are released, the license, copyright information, and terms of use in the package should be provided. For popular datasets, \url{paperswithcode.com/datasets} has curated licenses for some datasets. Their licensing guide can help determine the license of a dataset.
        \item For existing datasets that are re-packaged, both the original license and the license of the derived asset (if it has changed) should be provided.
        \item If this information is not available online, the authors are encouraged to reach out to the asset's creators.
    \end{itemize}

\item {\bf New assets}
    \item[] Question: Are new assets introduced in the paper well documented and is the documentation provided alongside the assets?
    \item[] Answer: \answerNA{} 
    \item[] Justification: The current submission does not release new datasets, model checkpoints, or public software artifacts. Therefore, documentation for newly released assets is not applicable.
    \item[] Guidelines:
    \begin{itemize}
        \item The answer \answerNA{} means that the paper does not release new assets.
        \item Researchers should communicate the details of the dataset\slash code\slash model as part of their submissions via structured templates. This includes details about training, license, limitations, etc. 
        \item The paper should discuss whether and how consent was obtained from people whose asset is used.
        \item At submission time, remember to anonymize your assets (if applicable). You can either create an anonymized URL or include an anonymized zip file.
    \end{itemize}

\item {\bf Crowdsourcing and research with human subjects}
    \item[] Question: For crowdsourcing experiments and research with human subjects, does the paper include the full text of instructions given to participants and screenshots, if applicable, as well as details about compensation (if any)? 
    \item[] Answer: \answerNA{} 
    \item[] Justification: The paper does not involve crowdsourcing, user studies, or experiments with human subjects.
    \item[] Guidelines:
    \begin{itemize}
        \item The answer \answerNA{} means that the paper does not involve crowdsourcing nor research with human subjects.
        \item Including this information in the supplemental material is fine, but if the main contribution of the paper involves human subjects, then as much detail as possible should be included in the main paper. 
        \item According to the NeurIPS Code of Ethics, workers involved in data collection, curation, or other labor should be paid at least the minimum wage in the country of the data collector. 
    \end{itemize}

\item {\bf Institutional review board (IRB) approvals or equivalent for research with human subjects}
    \item[] Question: Does the paper describe potential risks incurred by study participants, whether such risks were disclosed to the subjects, and whether Institutional Review Board (IRB) approvals (or an equivalent approval/review based on the requirements of your country or institution) were obtained?
    \item[] Answer: \answerNA{} 
    \item[] Justification: The paper does not involve human-subject research, so IRB approval or equivalent review is not applicable.
    \item[] Guidelines:
    \begin{itemize}
        \item The answer \answerNA{} means that the paper does not involve crowdsourcing nor research with human subjects.
        \item Depending on the country in which research is conducted, IRB approval (or equivalent) may be required for any human subjects research. If you obtained IRB approval, you should clearly state this in the paper. 
        \item We recognize that the procedures for this may vary significantly between institutions and locations, and we expect authors to adhere to the NeurIPS Code of Ethics and the guidelines for their institution. 
        \item For initial submissions, do not include any information that would break anonymity (if applicable), such as the institution conducting the review.
    \end{itemize}

\item {\bf Declaration of LLM usage}
    \item[] Question: Does the paper describe the usage of LLMs if it is an important, original, or non-standard component of the core methods in this research? Note that if the LLM is used only for writing, editing, or formatting purposes and does \emph{not} impact the core methodology, scientific rigor, or originality of the research, declaration is not required.
    \item[] Answer: \answerNA{} 
    \item[] Justification: The core method development in this research does not involve LLMs as any important, original, or non-standard components.
    \item[] Guidelines:
    \begin{itemize}
        \item The answer \answerNA{} means that the core method development in this research does not involve LLMs as any important, original, or non-standard components.
        \item Please refer to our LLM policy in the NeurIPS handbook for what should or should not be described.
    \end{itemize}

\end{enumerate}

\end{document}